%% file: main.tex
\documentclass[10pt,twocolumn,letterpaper]{article}

\usepackage{multirow}
\usepackage[table]{xcolor}
\usepackage{bm}

\usepackage[pagenumbers]{cvpr} 

\usepackage[accsupp]{axessibility} 


\newlength{\itemwidth} %

\definecolor{cvprblue}{rgb}{0.21,0.49,0.74}
\definecolor{tabfirst}{rgb}{1, 0.7, 0.7} 
\definecolor{tabsecond}{rgb}{1, 0.85, 0.7} 
\definecolor{tabthird}{rgb}{1, 1, 0.7} 

\usepackage[pagebackref,breaklinks,colorlinks,citecolor=cvprblue]{hyperref}

\usepackage[capitalize]{cleveref}
\crefname{section}{Sec.}{Secs.}
\Crefname{section}{Section}{Sections}
\Crefname{table}{Table}{Tables}
\crefname{table}{Tab.}{Tabs.}

\newcommand{\Fref}[1]{Fig.~\ref{#1}}
\newcommand{\Tref}[1]{Table~\ref{#1}}

\def\paperID{5244} 
\def\confName{CVPR}
\def\confYear{2024}

\title{Entity-NeRF: Detecting and Removing Moving Entities in Urban Scenes}

\author{
    Takashi Otonari\textsuperscript{1} \quad Satoshi Ikehata\textsuperscript{2,3,1} \quad Kiyoharu Aizawa\textsuperscript{1} \\
$^{1}$The University of Tokyo \quad ${^2}$National Institute of Informatics (NII) \quad ${^3}$Tokyo Institute of Technology\\
    {\tt\small \{otonari,aizawa\}@hal.t.u-tokyo.ac.jp \quad}
    {\tt\small 
    {sikehata@nii.ac.jp}}
}

\begin{document}

\twocolumn[{%
	\renewcommand\twocolumn[1][]{#1}%
	\maketitle
	\vspace{-2em}
    \centering
    \setlength{\tabcolsep}{0.02cm}
    \setlength{\itemwidth}{2.4cm}
    \renewcommand{\arraystretch}{0.5}
    \hspace*{-\tabcolsep}\small\begin{tabular}{ccccccc}
            \includegraphics[width=\itemwidth]{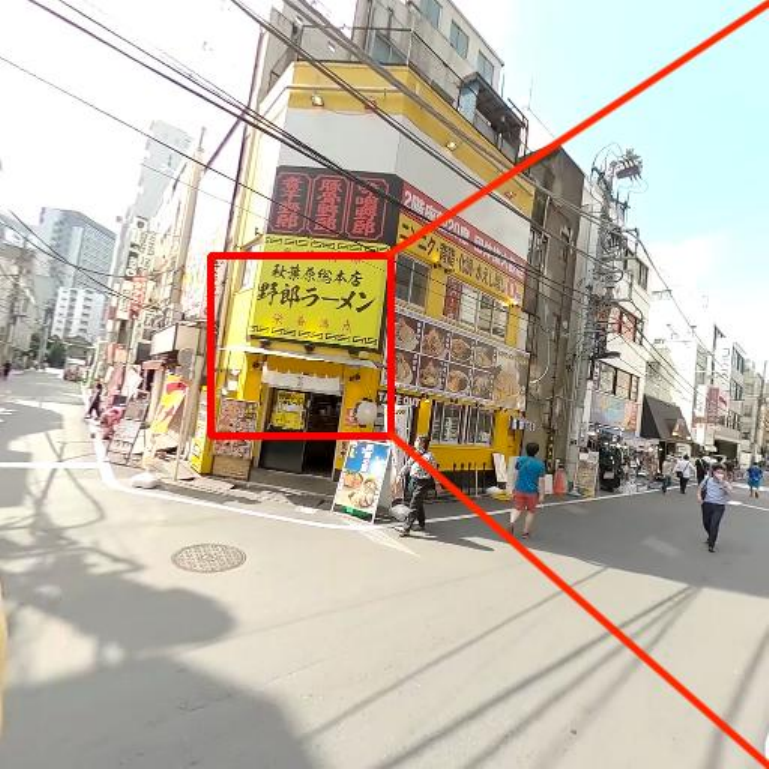} &
            \fboxsep=0pt\fcolorbox{red}{white}{\includegraphics[width=\itemwidth]{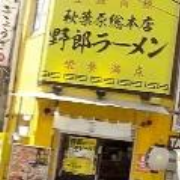}} &
            \includegraphics[width=\itemwidth]{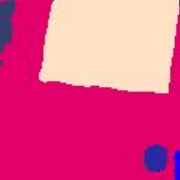} &
            \fboxsep=0pt\fbox{\includegraphics[width=\itemwidth]{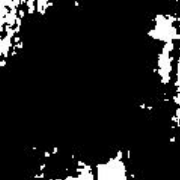}} &
            \fboxsep=0pt\fcolorbox{red}{white}{\includegraphics[width=\itemwidth]{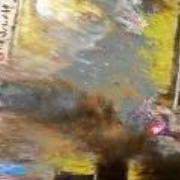}} &
            \fboxsep=0pt\fbox{\includegraphics[width=\itemwidth]{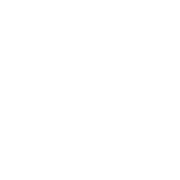}} &
            \fboxsep=0pt\fcolorbox{red}{white}{\includegraphics[width=\itemwidth]{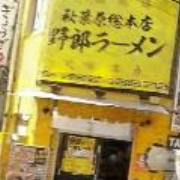}} \\
            \includegraphics[width=\itemwidth]{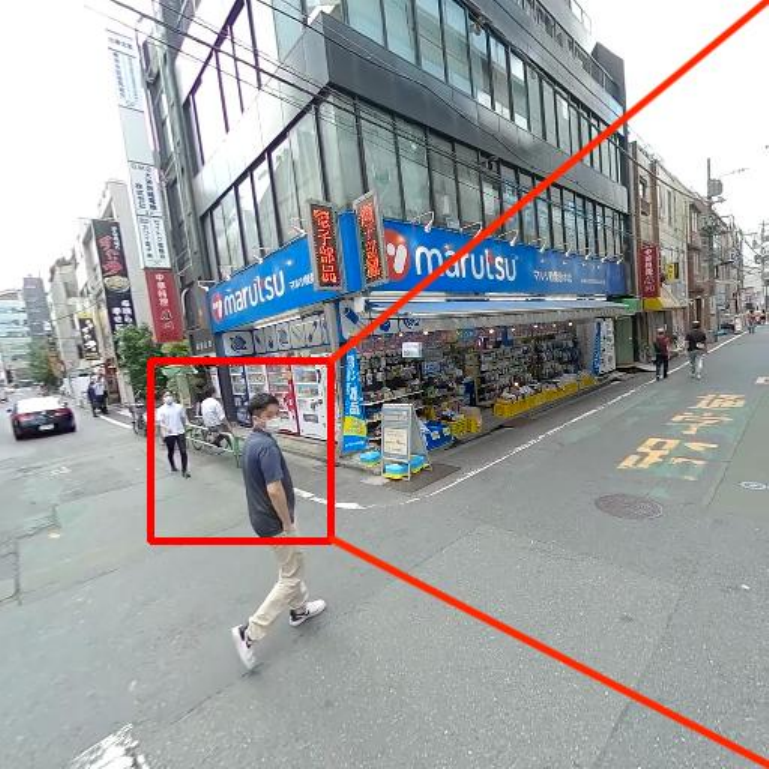} &
            \fboxsep=0pt\fcolorbox{red}{white}{\includegraphics[width=\itemwidth]{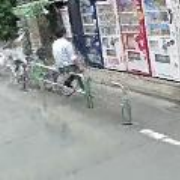}} &
            \includegraphics[width=\itemwidth]{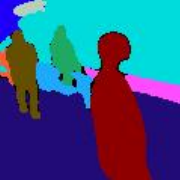} &
            \fboxsep=0pt\fbox{\includegraphics[width=\itemwidth]{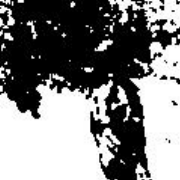}} &
            \fboxsep=0pt\fcolorbox{red}{white}{\includegraphics[width=\itemwidth]{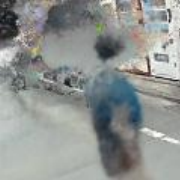}} &
            \fboxsep=0pt\fbox{\includegraphics[width=\itemwidth]{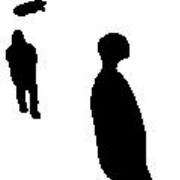}} &
            \fboxsep=0pt\fcolorbox{red}{white}{\includegraphics[width=\itemwidth]{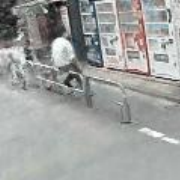}} \\
            \vspace{0.2em}
            Original image &
            Rendered &
            Entity Seg.~\cite{qilu2023high} &
            RobustNeRF~\cite{robustnerf} &
            RobustNeRF~\cite{robustnerf} &
            Entity-NeRF &
            Entity-NeRF \\
            & ground-truth & & weights & & weights & \\
        \\
    \end{tabular}\vspace{-1.5em}
    \captionof{figure}{In urban scenes, statistical approach~\cite{robustnerf} mistakes complex backgrounds for moving objects (top) and fails to remove small moving objects (bottom). On the other hand, Entity-NeRF can reconstruct complex backgrounds and remove small moving objects.}
    \label{fig:teaser}
    \vspace{1.3em}
}]

\begin{abstract}
Recent advancements in the study of Neural Radiance Fields (NeRF) for dynamic scenes often involve explicit modeling of scene dynamics. However, this approach faces challenges in modeling scene dynamics in urban environments, where moving objects of various categories and scales are present. In such settings, it becomes crucial to effectively eliminate moving objects to accurately reconstruct static backgrounds. Our research introduces an innovative method, termed here as Entity-NeRF, which combines the strengths of knowledge-based and statistical strategies. This approach utilizes entity-wise statistics, leveraging entity segmentation and stationary entity classification through thing/stuff segmentation. To assess our methodology, we created an urban scene dataset masked with moving objects. Our comprehensive experiments demonstrate that Entity-NeRF notably outperforms existing techniques in removing moving objects and reconstructing static urban backgrounds, both quantitatively and qualitatively.~\footnote{Our project page is available at \href{https://otonari726.github.io/entitynerf/}{https://otonari726.github.io/entitynerf/}}
\end{abstract}

\section{Introduction}
\label{sec:intro}
Novel view synthesis is rapidly evolving, which enables the creation of new visual content such as the immersive views found in Google Maps and the free-viewpoint visualizations in sports broadcasts. However, a key innovation in this field, Neural Radiance Fields (NeRF)~\cite{nerf}, faces challenges when dealing with urban scenes whose complexity is inherently high due to a large number of dynamic elements present, such as moving vehicles, pedestrians, changing lighting conditions, and varying shadows. The ability to accurately render and reconstruct such scenes is crucial for several applications including autonomous navigation, surveillance, and virtual urban exploration among others. 

The challenge of handling dynamic scenes has been a notable point of extension within the domain of Neural Radiance Fields, and two major approaches prevail.  The first involves explicit modeling of scene dynamics, which concurrently encodes both static and dynamic information, exemplified by methods such as D-NeRF~\cite{pumarola2020d}, HyperNeRF~\cite{park2021hypernerf}, and RoDynRF~\cite{liu2023robust}. The second approach adopts a more statistical perspective, treating scene dynamics as outliers in relation to the static elements~\cite{robustnerf}. The elimination of dynamic elements in the scene contributes to a reduction in clutter, enhances the comprehension of the scene, and improves visual fidelity. 

Despite the progress in research on novel-view synthesis of dynamic scenes, to our knowledge, there is no effective method for unbounded scenes like urban environments, where a multitude of moving objects of various categories and scales, such as people, cars, and bicycles, coexist. For instance, current methods in the former category often target specific objects, deal with a minority of moving objects within a scene, and are restricted to narrow scene boundaries. Moreover, while the latter approach can handle multiple objects simultaneously, it solely relies on the statistics of reconstruction errors for outlier separation and does not function effectively when dynamic objects vary in scale or when the background is complex, as shown in \Fref{fig:teaser}.

In this work, we address the task of learning static NeRFs of dynamic urban scenes. To identify a multitude of moving objects of various categories and scales, we propose a hybrid method that integrates the strengths of both knowledge-based and statistical approaches with Entity-wise Average of Residual Ranks (EARR) and stationary entity classification. EARR identifies distractors by entity-wise statistics based on entity segmentation~\cite{qilu2023high}. In addition, the stationary entity classification using the thing/stuff segmentation~\cite{ade20k,xie2021segformer} enables more efficient learning by incorporating complex backgrounds such as building from the early stages of learning.

To evaluate our proposed method, we annotated moving objects in three videos captured in urban scenes with challenging settings and rendered images using a static NeRF which removed the masked moving objects. Using the overall Peak Signal-to-Noise Ratio (PSNR) of the image to measure whether moving objects, which constitute only a small portion of the image, have been appropriately removed is challenging. Therefore, we evaluate the moving object removal  (foreground PSNR) and the static background reconstruction (background PSNR) separately. Our experiments demonstrate that our method effectively removes moving objects and reconstructs static backgrounds in urban scenes, while still maintaining accuracy on existing datasets.

\section{Related Works}
\label{sec:related_works}
\begin{figure*}[!t]
    \centering
    \includegraphics[width=\linewidth]{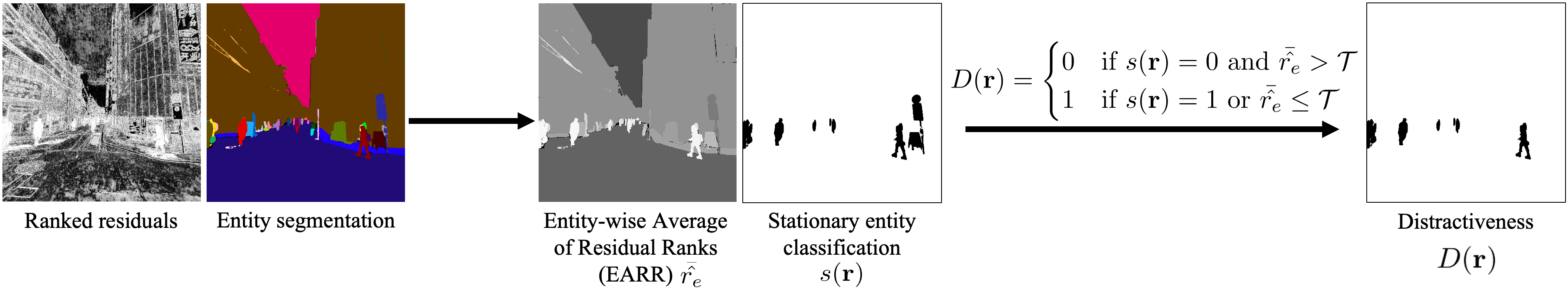}
    \vspace{-0.75cm}\caption{\textbf{Overview of our Entity-NeRF pipeline.} $D(\mathbf{r}) = 0$ if Entity-wise Average of Residual Ranks (\ref{subsec:entity-wise-loss}) of the entities labeled `thing' in the stationary entity classification (\ref{subsec:neural-weight-function}) is greater than a threshold value $\mathcal{T}$. The `thing' label for the stationary entity classification is given as $s(\mathbf{r})=0$ and the `stuff' label as $s(\mathbf{r})=1$.}
    \label{fig:entity-wise-loss-pipeline}
\end{figure*}

\subsection{NeRF on Dynamic Scenes}
NeRF~\cite{nerf} represents coordinate-based neural networks that predict the radiance from a specific view and opacity at any given 3-D coordinate. To render novel views of a scene via ray-tracing,  NeRF is trained by minimizing the difference between each pixel’s rendered and observed colors given calibrated multi-view images of the scene.

In the original NeRF, handling dynamic scenes is challenging due to the inherent assumption that the entire scene remains static. To address this issue, subsequent research has proposed methods that either explicitly learn scene dynamics by category-specific methods~\cite{peng2021neural,saito2019pifu,weng2022humannerf,Wang2023ClothedHumanCap,liu2023hosnerf,yu2023monohuman,SHERF,liu2021neural,peng2021animatable,su2021anerf,Gafni_2021_CVPR,Raj_2021_CVPR}, detection~\cite{Ost_2021_CVPR,KunduCVPR2022PNF}, deformation~\cite{park2021nerfies,park2021hypernerf,pumarola2020d,tretschk2021non,wang2021neural,Ma_2023_ICCV,Wang_2023_CVPR,yu2023dylin,yan2023nerf}, flow~\cite{li2020neural,fsnerf,gao2021dynamic,du2021neural,yoon2020novel}, multiple synchronized videos~\cite{li2022neural,wang2022fourier,zhang2021editable}, depth-based approaches~\cite{xian2021space}, or treat moving objects as outliers in a robust approach~\cite{robustnerf}.
\\
\noindent \textbf{Detection-based Approach:}
Neural Scene Graphs~\cite{Ost_2021_CVPR} and Panoptic Neural Fields~\cite{KunduCVPR2022PNF} provide a structured approach to explicitly detecting and modeling individual dynamic objects within dynamic scenes. While these methods facilitate object-level manipulation, moving object detection is hindered by occlusions, diverse object types, and scales in the urban environment.
\\\\
\noindent \textbf{Deformation-based Approach:}
Without object detection, some methods such as D-NeRF~\cite{pumarola2020d}, Nerfies~\cite{park2021nerfies} and HyperNeRF~\cite{park2021hypernerf} represent scenes using a deformation field, mapping observations to neighboring frames or a canonical scene. However, they are limited to small-motion, object-centric scenes due to challenges in representing entire sequences with a single canonical voxel.\\

In recent efforts, $\mathrm{D^2}$NeRF~\cite{d2nerf} separates moving objects, static backgrounds, and shadows into three fields using regularization. DynIBaR~\cite{li2023dynibar} aggregates multi-view image features in a motion-adjusted ray space, while RoDynRF~\cite{liu2023robust} uses a time-dependent MLP and single-view depth priors. FSDNeRF~\cite{fsdnerf} uses data-driven optical flow for backward deformation computation to handle rapid motion.

In urban settings, accurate optical flow estimation faces challenges due to numerous cluttered objects of varying scales, which often result in incorrect dynamics modeling. Moreover, deformation-based methods struggle with frames having large temporal steps, constraining their urban modeling use from a discrete set of multi-view images.
\\\\
\textbf{Robust Approach:}
Relatively little attention has been paid to removing non-static elements from discrete multi-view images rather than a continuous video stream. One straightforward approach is to segment and ignore pixels during training that are likely to be transient objects~\cite{rematas2022urf, tancik2022blocknerf}, for example, by applying a data-driven segmentation model. However, removing objects based on object semantics is risky since semantic segmentation is far from perfect and runs the risk of erroneously removing static objects that are typically mobile (e.g., cars, pedestrians).

We can also use a robust estimator. RobustNeRF~\cite{robustnerf} has proposed a purely statistical approach to remove moving elements as outliers by analyzing the patch-wise statistics of reconstruction errors. By formulating training as a form of iterative reweighted least squares, this method can robustly separate inliers and outliers, which are not limited to specific predefined categories, from photo collections rather than videos. While effective, this method fixes the hyperparameters such as patch size, which becomes problematic when applied to urban scenes where there is a variety of moving objects of different types and scales.

\subsection{NeRF on Unbounded Scenes}
The original NeRF struggles with unbounded scenes due to sparse rays at greater distances. Adaptations like NeRF++\cite{kaizhang2020}, which introduced inverted sphere parameterization, F2-NeRF\cite{wang2023f2nerf} with its perspective warping, and scene contraction approaches by MERF~\cite{merf} and Mip-NeRF 360~\cite{barron2022mipnerf360}, have been developed to address this. Recent developments, Zip-NeRF~\cite{barron2023zipnerf} and Nerfacto~\cite{nerfstudio}, further refine these methods for unbounded environments.

For large city-scale scenes, Block-NeRF~\cite{tancik2022blocknerf}, Mega-NeRF~\cite{Turki_2022_CVPR}, and SUDS~\cite{turki2023suds} segment scenes into blocks, applying NeRF within each. SUDS notably integrates additional data types like LiDAR for dynamic city-scale scenes.

\subsection{Entity Segmentation}
Entity segmentation is a new class of image segmentation tasks aiming to segment all entities in an image without predicting their semantic or instance labels~\cite{qi2022open, shen2021high, qi2022cassl, qilu2023high}. Eliminating the need for class labels is helpful for many practical applications, such as image manipulation and editing, where the quality of segmentation masks is crucial, but class labels are less important.

Recently, Qi~\etal~\cite{qilu2023high} presented a large-scale entity segmentation dataset and proposed CropFormer, a Transformer-based entity segmentation method. In our research, we utilize this result for training NeRF on urban dynamic scenes and demonstrate that Entity-wise Avarage of Residual Ranks (EARR) overcomes most issues found in patch-based counterparts in RobustNeRF~\cite{robustnerf}.

\section{Preliminaries}
\label{sec:preliminary}
\subsection{Problem Statement}
In our context, there is no need to explicitly model moving objects. Therefore, our goal is to simply detect and segment out moving objects in the scene as distractors during the basic training pipeline of arbitrary NeRF models~\cite{nerf,barron2021mipnerf}. More concretely, we want to label the distractiveness $D(\mathbf{r})$ to each ray $\mathbf{r}$ and reflect labels in photometric reconstruction losses in training NeRF as 
\begin{equation}
\begin{aligned}
\mathcal{L}_{\mathbf{r}} &= D(\mathbf{r}) \cdot \epsilon(\mathbf{r}) \\
\epsilon(\mathbf{r}) &= \| C_{\mathrm{gt}}(\mathbf{r}) - C_{\mathrm{pred}}(\mathbf{r}) \|_2^2
\end{aligned}
\end{equation}

where

\begin{equation}
D(\mathbf{r}) = 
\begin{cases} 
0 & \text{if } \mathbf{r} \text{ passes through a distractor,} \\
1 & \text{otherwise.}
\end{cases}
\end{equation}

Here, $C_\mathrm{pred}$ and $C_\mathrm{gt}$ are rendered and observed pixel colors and $\epsilon(\mathbf{r})$ is the $\ell_2$ residual of them. While the task is formulated in a simple form, predicting $D(\textbf{r})$ in urban scenes poses several fundamental challenges as detailed below.

\begin{figure*}[t]
    \centering
    \setlength{\tabcolsep}{0.02cm}
    \setlength{\itemwidth}{3.4cm}
    \renewcommand{\arraystretch}{0.5}
    \hspace*{-\tabcolsep}\small\begin{tabular}{ccccc}
            \includegraphics[width=\itemwidth]{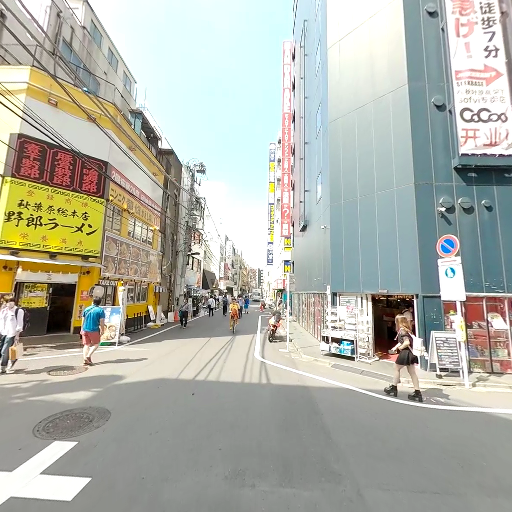} &
            \includegraphics[width=\itemwidth]{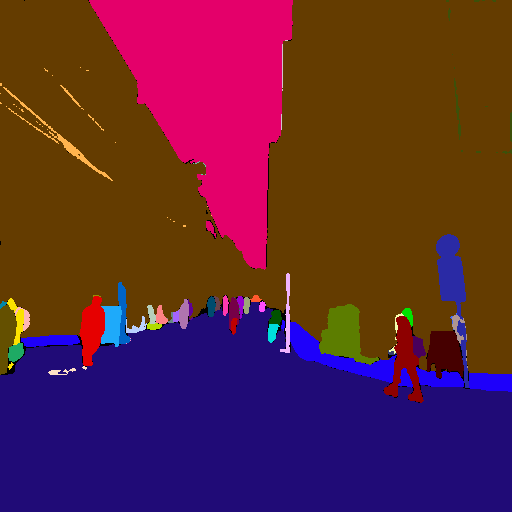} &
            \fboxsep=0pt\fbox{\includegraphics[width=\itemwidth]{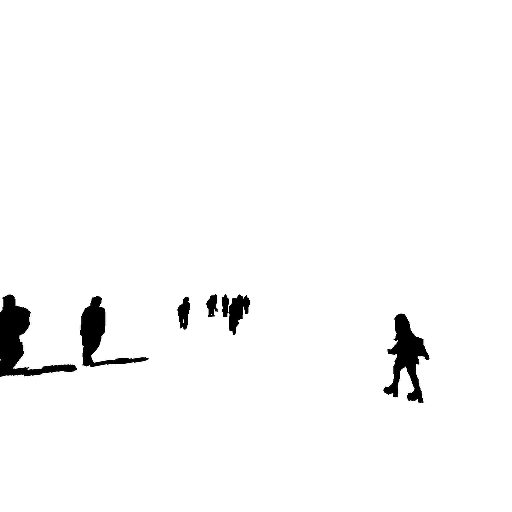}} &
            \fboxsep=0pt\fbox{\includegraphics[width=\itemwidth]{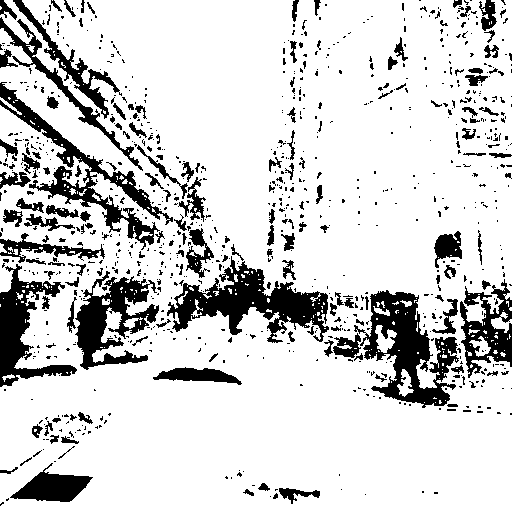}} &
            \fboxsep=0pt\fbox{\includegraphics[width=\itemwidth]{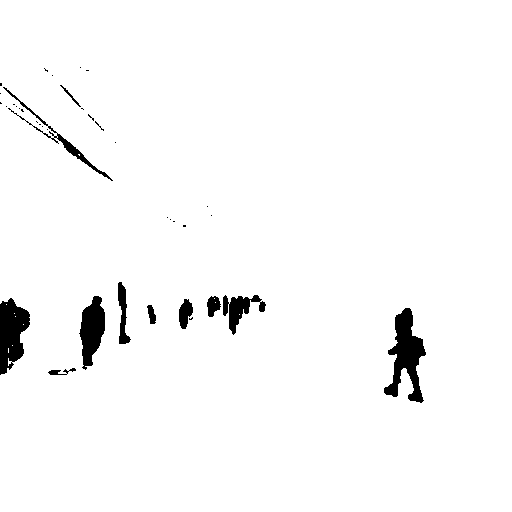}} \\
            \includegraphics[width=\itemwidth]{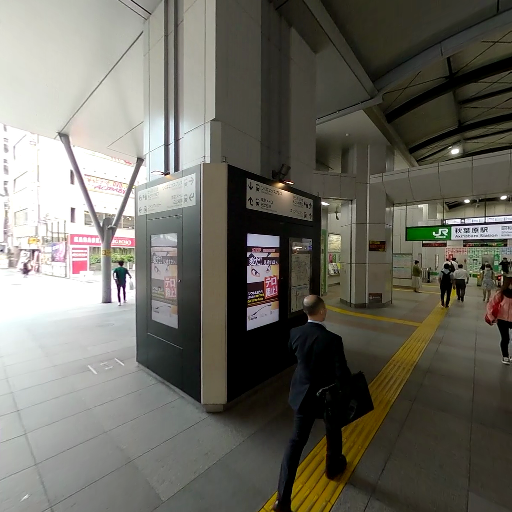} &
            \includegraphics[width=\itemwidth]{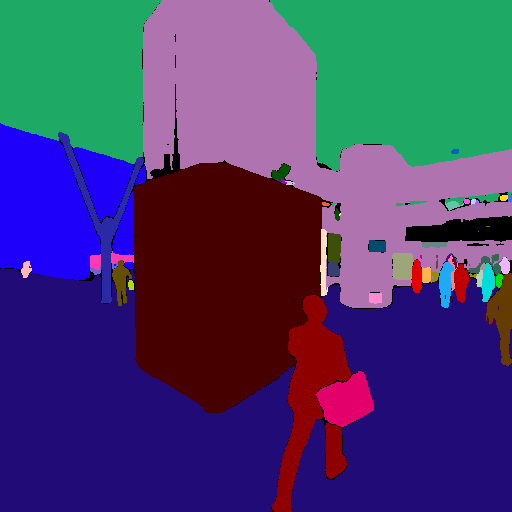} &
            \fboxsep=0pt\fbox{\includegraphics[width=\itemwidth]{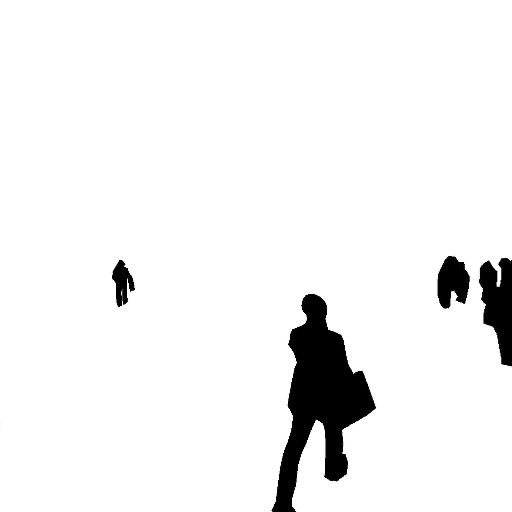}} &
            \fboxsep=0pt\fbox{\includegraphics[width=\itemwidth]{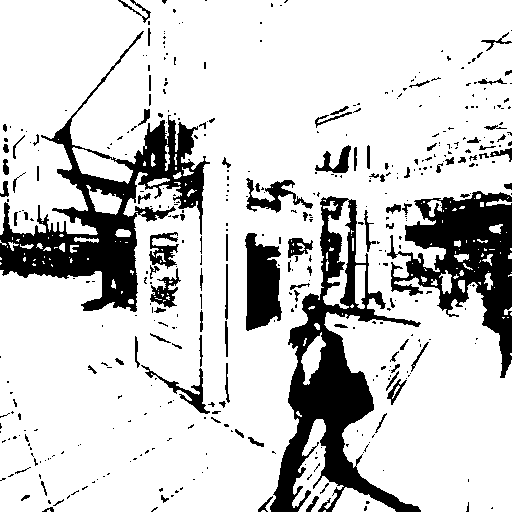}} &
            \fboxsep=0pt\fbox{\includegraphics[width=\itemwidth]{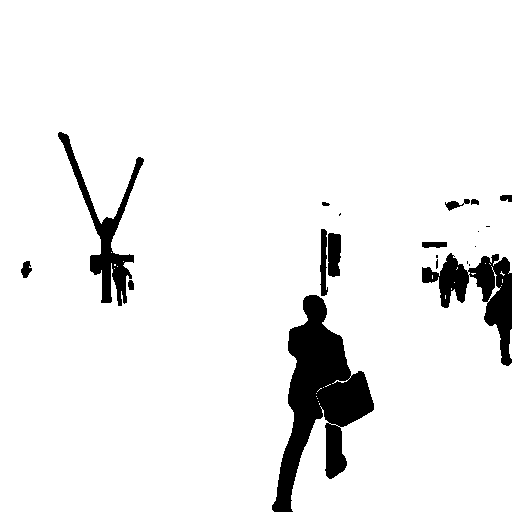}} \\\vspace{0.2em}
            Original image &
            Entity segmentation~\cite{qilu2023high} &
            Mask for moving objects &
            RobustNeRF~\cite{robustnerf} &
            EARR \\
        \\
    \end{tabular}\vspace{-1.5em}
  \caption{\textbf{\(\bm{D(}\mathbf{r}\bm{)}\) of RobustNeRF~\cite{robustnerf} and our Entity-wise Average of Residual Ranks (EARR) at the end of training.} Our EARR can more efficiently incorporate the background into learning.}
  \label{fig:vis_weight}
\end{figure*}

\subsection{Challenges in Urban Scenes}
Firstly, prior knowledge of scene semantics is often incorporated to specify moving objects (\eg,~\cite{rematas2022urf, tancik2022blocknerf}), as it provides accurate masks along the object's contours to a certain extent. However, relying solely on per-image scene semantics to specify distractors is insufficient. Urban environments feature a wide variety of moving objects, ranging from people, and vehicles, to minor elements like roadside cans. These objects are not always covered by standard semantic segmentation classes. Furthermore, even within common classes such as vehicles and pedestrians, they cannot always be identified as distractors based on class alone, as a parked car, for instance, should not be classified as a distractor.

In addition, a purely statistical approach (\eg,~\cite{robustnerf}) may not always successfully identify distractors in urban scenes. For instance, RobustNeRF~\cite{robustnerf} assigns inlier/outlier labels to each non-overlapping $8\times 8$ patch, considering only the statistics of reconstruction errors within $16\times 16$ neighboring pixels. Due to this heuristic, RobustNeRF is only effective when the background is relatively simple when the reconstruction errors decrease rapidly, and the size of distractors is significantly larger than the predefined neighbor system. However, in most complex urban scenes, the varying distances to moving objects and the diverse coverage of field-of-view may hinder the statistical method's ability to distinguish between inliers and outliers effectively.

To overcome these limitations, we propose a hybrid method that integrates the strengths of both knowledge-based and statistical approaches. Specifically, we leverage the knowledge-based entity segmentation method's ability to generate accurate object contours and semantic segmentation method to identify non-moving objects, while incorporating the capability of the statistical method in adaptively handling varied scene dynamics and object movements. This synergy aims to create a more robust and versatile system for identifying distractors in urban environments, addressing the limitations of each method when used in isolation.

\section{Method}
\label{sec:method}
In this section, we present our hybrid approach combining knowledge-based and statistical methods to identify moving distractors of varying sizes in urban scenes. Concretely, we introduce the method of distractor labeling using Entity-wise Average of Residual Ranks (EARR) (\ref{subsec:entity-wise-loss}), utilizing both data-driven segmentation networks and entity-wise statistics of reconstruction losses. As the statistics of reconstruction losses become highly unstable in complex background areas, we incorporate knowledge of scene semantics to identify the non-moving stuff, such as buildings (\ref{subsec:neural-weight-function}). The overall pipeline is illustrated in \Fref{fig:entity-wise-loss-pipeline}.

\subsection{Entity-wise Average of Residual Ranks (EARR)}
\label{subsec:entity-wise-loss}
A pre-trained entity segmentation network \cite{qilu2023high} provides high-quality segmentation for objects in real-world scenes including urban scenes. This quality is maintained regardless of the objects' semantics or sizes in the image. Although it's not possible to determine if an entity is moving based on the segmentation result alone, we can assume that there is consistency in the distractor label across pixels within the same entity. This observation leads to a departure from conventional methods. Instead of assigning moving distractor labels to rays that intersect each pixel \cite{tancik2022blocknerf} or each patch \cite{robustnerf}, our approach labels individual entities.

To determine whether each entity is moving or not, we utilize the statistics of reconstruction loss in each entity. Similar to RobustNeRF~\cite{robustnerf}, we follow the principle that rays passing through distractors lead to a lack of consistency across multiple viewpoints, resulting in larger reconstruction loss.

To clarify, we denote $\epsilon(i)$ as the $\ell_2$ residual between rendered and observed colors of a ray passing through the $i$-th pixel. Considering $N$ pixels (\ie, $N$ rays) in a batch, we define a rank function $R(\epsilon(i))$ that inputs $\epsilon(i)$ and outputs an ordered rank, with the largest residual assigned $N$ and the smallest assigned 1 among the $N$ pixels. Then, the normalized residual rank $\hat{r}(i)$ is calculated by normalizing these ranks to the [0, 1] range and used for the distractor labeling in the following steps. The use of normalized residual rank instead of raw residual values is justified because, during the initial stages of training, the residuals tend to be large. Relying solely on the raw residual values for decision-making can lead to excessive false detection of distractors. In contrast, by employing a rank function, it is possible to exclude only a specific proportion of samples with large residual values, while ensuring that all other samples are included in the training process.

The normalized residual rank for each ray tends to be higher when the ray passes through a moving object, due to an increase in the residual. However, even in static scenes, the rank can become high if the ray passes through complex backgrounds or geometric shapes. Therefore, instead of using a single ray as the basis for decision-making, we gather statistics on the Rank of each entity to use as a basis for labeling. 

However, since the shape and size of entities vary greatly, it is not practical to sample rays passing through all pixels of each entity during training. To address this, we sample a patch of size $k \times k$ pixels (\ie, $k$ should be sufficiently large and we choose $64$ in our implementation), cluster the labels of entities within it, and then calculate statistical measures for each cluster. Specifically, for a set of pixels corresponding to an entity ID ($e$) within a patch, designated as $S(e)$, we calculate its average as follows.
\begin{align}
    \Bar{\hat{r_e}} = \frac{\Sigma_{i\in S(e)} \hat{r}(i)}{\left|S(e)\right|}~. \label{eq:mean_rank}
\end{align}
where, the number of elements in each entity is denoted as $\left|S(e)\right|$. This process is repeated for all entities in the patch. If the average exceeds a certain threshold $\mathcal{T}$, we assign $D(\mathbf{r})=0$ and otherwise $D(\mathbf{r})=1$ to all rays passing through pixels corresponding to that entity ID. The choice of $\mathcal{T}$ is crucial and its choice will be discussed later.

Despite its simplicity, our approach, which combines knowledge-based entity segmentation results and statistics of residual ranks, functions much more robustly than methods based solely on statistics. As depicted in~\Fref{fig:vis_weight}, our proposed method has been shown to accurately detect nearly all moving objects without excessively excluding inliers. This is in contrast to RobustNeRF~\cite{robustnerf}, a purely statistical approach, which is prone to excessively identify distractors.

\subsection{Cooperative Stationary Entity Classification}
\label{subsec:neural-weight-function}
In the early stages of training, all residuals are high, leading to low reliability of residual ranks and their statistical measures. This is especially true in urban scenes where backgrounds contain numerous elements, such as traffic signs and complex building structures. These elements make NeRF training difficult, resulting in many inliers being included in samples above the rank threshold. These inliers could potentially be excluded from training during large training steps. To address this issue, for entities of classes such as buildings, sky, and roads in urban scenes, which are certainly stationary, we attempt to assign a value of 1 to \(D(\mathbf{r})\) to ensure their inclusion in the learning process regardless of their residual ranks.

To implement this, we train a stationary entity classification network, which is an MLP with three linear layers and one classification layer. This network identifies whether each entity belongs to a class of stationary objects. The input to this network is a feature vector calculated for each entity, and the output is a class label of the entity, either `stuff' or `thing'. `stuff' and `thing' are defined in ADE20K~\cite{ade20k} as non-accumulative and accumulative objects, respectively. Specifically, all movable object classes are included in `thing', and we can safely include `stuff' entities in the training. Feature vectors for individual entities are computed by averaging pixel-wise features within each entity. Concretely, feature maps are extracted from an image by applying pre-trained SAM~\cite{sam} and DINOv2~\cite{oquab2023dinov2} encoders, individually.

For training this network, we also adopt a cooperative approach combining prior knowledge and statistics. Specifically, instead of training the stationary entity classification network entirely on ADE20K, we adapt the network for individual scenes. This adaptation is done by continuously fine-tuning the MLP using pixels classified as stationary (\(D(\mathbf{r})=1\)) during the training, based on the ranked residuals described above. Concretely, we first train SegFormer~\cite{xie2021segformer} on ADE20K to output `stuff'/`thing' labels, then apply it to each scene in which NeRF is trained to generate initial pseudo ground truth labels for stationary entities specific to each scene. Note that the labels for each entity are determined based on the voting of pixel-wise labels for each entity. The four-layer MLP is then pre-trained based on these pseudo ground truth labels. Every 100 steps of NeRF training, the MLP is fine-tuned using the entities that have been determined as \(D(\mathbf{r})=1\) based on the ranked residuals. Entities for which the stationary entity classification network assigns `stuff' labels are consistently included in the NeRF training, whereas entities assigned `thing' labels are trained solely based on ranked residuals.

\begin{table*}[t]
    \centering
    \small\begin{tabular}{c|c|rr|rrr}
    Model & Loss & \begin{tabular}
    {c}foreground\\PSNR$\uparrow$\end{tabular} & \begin{tabular}{c}background\\PSNR$\uparrow$\end{tabular} & PSNR$\uparrow$ & SSIM$\uparrow$ & LPIPS$\downarrow$ \\\hline
    \multirow{4}{*}{Nerfacto~\cite{nerfstudio}} & Mean-squared error (MSE) & 12.10 & \cellcolor{tabfirst}25.07 & \cellcolor{tabfirst}24.96 & \cellcolor{tabfirst}0.87 & \cellcolor{tabfirst}0.10 \\
    & RobustNeRF~\cite{robustnerf} & 17.63 & 21.74 & 23.19 & 0.84 & \cellcolor{tabsecond}0.12 \\
    & Entity-NeRF (only EARR) & \cellcolor{tabsecond}19.48 & 23.68 & 24.63 & 0.84 & 0.13 \\
    & \textbf{Entity-NeRF}  & \cellcolor{tabfirst}19.82 & \cellcolor{tabsecond}24.00 & \cellcolor{tabsecond}24.93 & \cellcolor{tabsecond}0.85 & \cellcolor{tabsecond}0.12 \\ \hline
    \multirow{4}{*}{Mip-NeRF 360~\cite{barron2022mipnerf360}} & Mean-squared error (MSE) & 11.40 & \cellcolor{tabfirst}27.36 & 24.22 & \cellcolor{tabfirst}0.88 & \cellcolor{tabfirst}0.13 \\
    & RobustNeRF~\cite{robustnerf} & 20.15 & 22.52 & 22.87 & 0.83 & 0.18 \\
    & Entity-NeRF (only EARR) & \cellcolor{tabsecond}20.20 & 25.49 & \cellcolor{tabsecond}25.21 & \cellcolor{tabsecond}0.85 & \cellcolor{tabsecond}0.14 \\
    & \textbf{Entity-NeRF} & \cellcolor{tabfirst}20.74 & \cellcolor{tabsecond}25.50 & \cellcolor{tabfirst}25.23 & 0.84 & 0.15 \\
    \end{tabular}
    \caption{\textbf{Quantitative comparison with RobustNeRF~\cite{robustnerf} using Mip-NeRF 360~\cite{barron2022mipnerf360} and Nerfacto~\cite{nerfstudio} on MovieMap Dataset.}}
    \label{table:experiment_realdata}
\end{table*}

\section{Results}
\label{sec:results}
\subsection{Implementation Details}
Our framework for labeling moving objects can be applied to all NeRF models that use photometric reconstruction loss similar to other robust approaches (\eg, RobustNeRF~\cite{robustnerf}). In this paper, we incorporated our method into both Mip-NeRF 360~\cite{barron2022mipnerf360} and Nerfacto~\cite{nerfstudio}. The implementation of each method is as follows. Note that we used the hyperparameter values from the original implementations. Mip-NeRF 360 trained on two Tesla A100 units and Nerfacto on one, using 16,384 samples during every iteration.

\begin{figure}[t]
    \centering
    \setlength{\tabcolsep}{0.02cm}
    \setlength{\itemwidth}{2.7cm}
    \renewcommand{\arraystretch}{0.5}
    \hspace*{-\tabcolsep}\begin{tabular}{ccc}
            \includegraphics[width=\itemwidth]{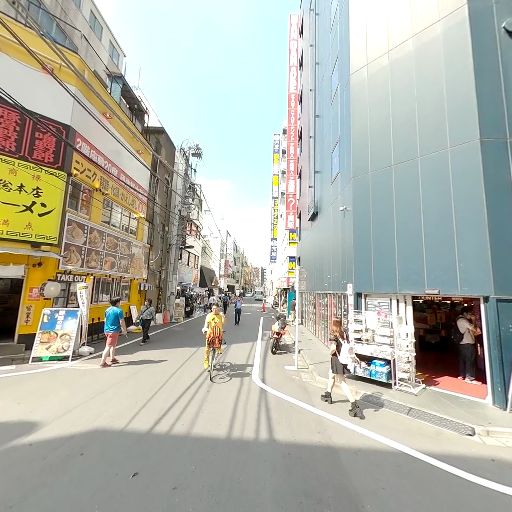} &
            \includegraphics[width=\itemwidth]{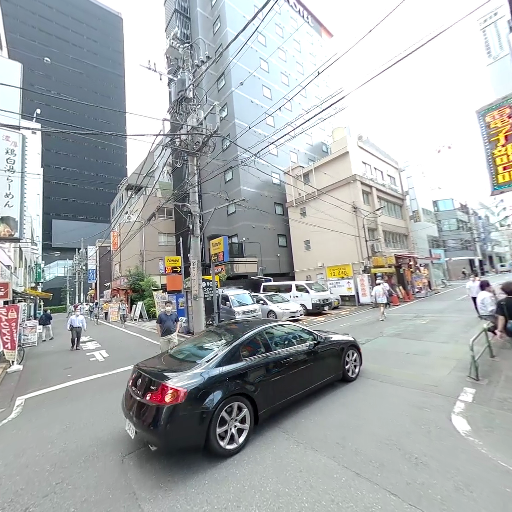} &
            \includegraphics[width=\itemwidth]{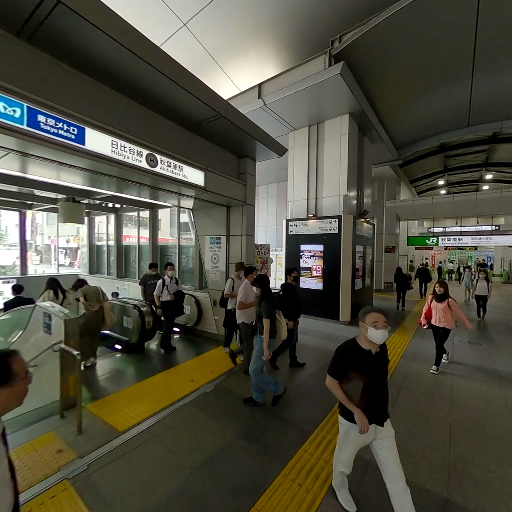} \\
            \fboxsep=0pt\fbox{\includegraphics[width=\itemwidth]{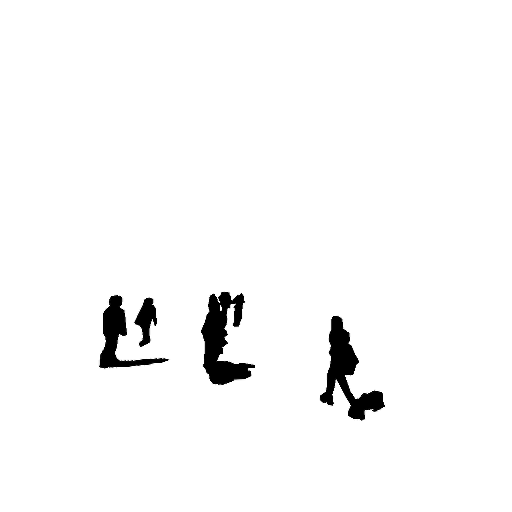}} &
            \fboxsep=0pt\fbox{\includegraphics[width=\itemwidth]{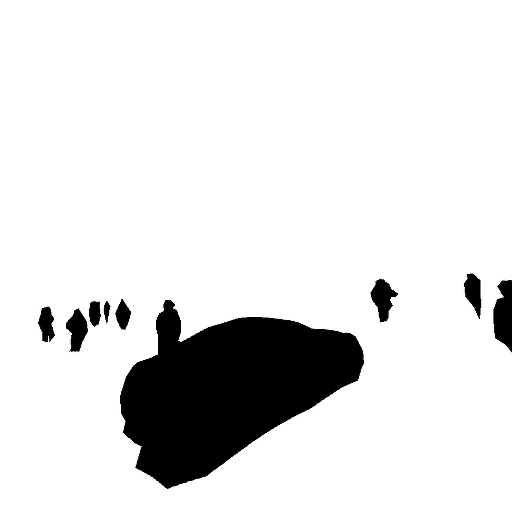}} &
            \fboxsep=0pt\fbox{\includegraphics[width=\itemwidth]{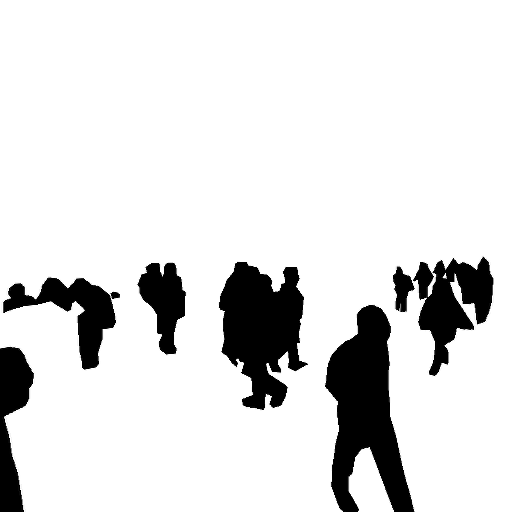}} \\        \\
    \end{tabular}\vspace{-1em}
  \vspace{-0.25cm}\caption{\textbf{MovieMap Dataset.} Only moving objects in the video are masked. Therefore, parked cars and stationary people are not masked.}
  \label{fig:vis_our_data}
\end{figure}

\noindent \textbf{Mip-NeRF 360}: Mip-NeRF 360 addresses challenges presented by unbounded scenes using a non-linear scene parameterization. We used the official implementation code from MultiNeRF~\cite{multinerf2022}, which contains an implementation of Mip-NeRF 360~\cite{barron2022mipnerf360} and RobustNeRF~\cite{robustnerf}. The models were trained for 250,000 iterations for each scene, taking approximately 24 hours.

\noindent \textbf{Nerfacto}: Nerfacto is implemented in NeRFStudio~\cite{nerfstudio}, which is a combination of various methods, rather than a single published work, that has proven effective in real-world applications. The models were trained for 30,000 iterations for each scene, taking approximately 30 minutes.

The most important hyperparameter in our method is the threshold parameter for the averaged residual rank, $\mathcal{T}$. If this value is too high, it includes too many outliers as inliers, and if too low, it excludes inliers as outliers. To prevent the inclusion of outliers excessively as inliers during training, we set the threshold value to $\mathcal{T}=0.8$. This decision is based on 78 manually annotated images from three scene images, with an average ratio of inliers being approximately 90.4\%. The impact of this value is also evaluated in the next chapter.

\subsection{Datasets}
The proposed method was quantitatively evaluated using two real-world datasets. The first dataset is an urban scene dataset (MovieMap Dataset), generated from 360\textdegree~videos from Movie Map~\cite{moviemap}, while the second comprises non-urban scenes published in~\cite{robustnerf} (RobustNeRF Dataset). Please see the supplementary for more details on RobustNeRF Dataset and quantitative comparisons.

\begin{figure}[!t]
    \centering
    \setlength{\tabcolsep}{0.02cm}
    \setlength{\itemwidth}{2.0cm}
    \renewcommand{\arraystretch}{0.5}
    \hspace*{-\tabcolsep}\small\begin{tabular}{cccc}
            \includegraphics[width=\itemwidth]{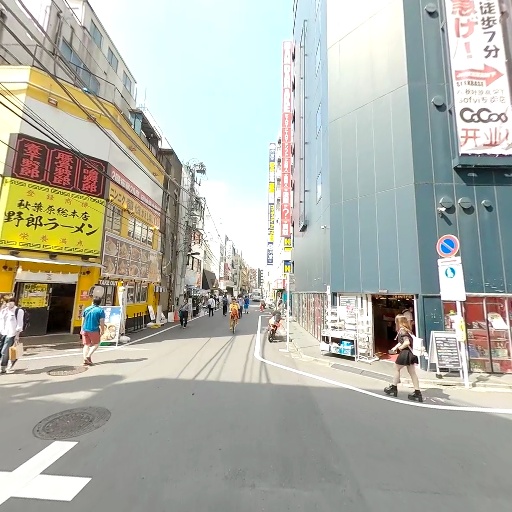} &
            \includegraphics[width=\itemwidth]{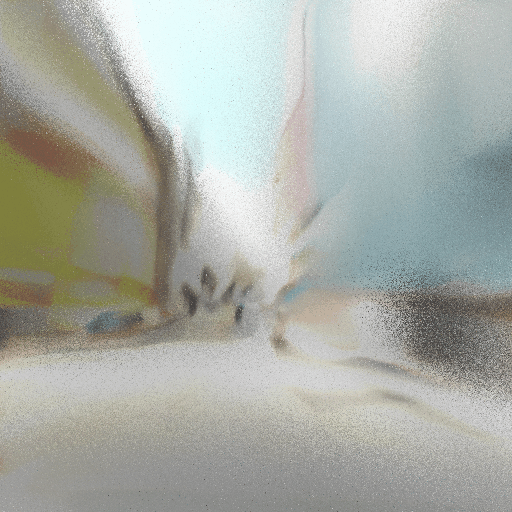} &
            \includegraphics[width=\itemwidth]{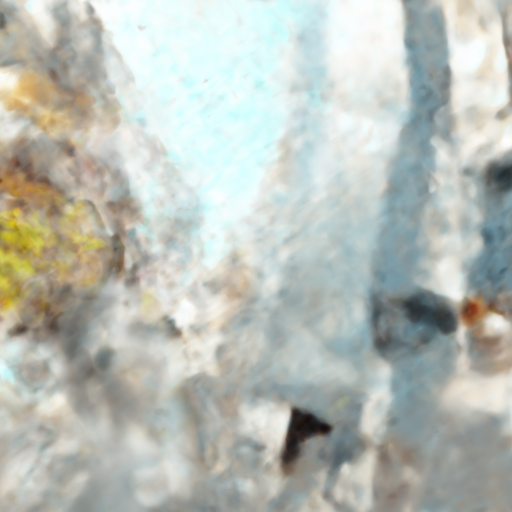} &
            \includegraphics[width=\itemwidth]{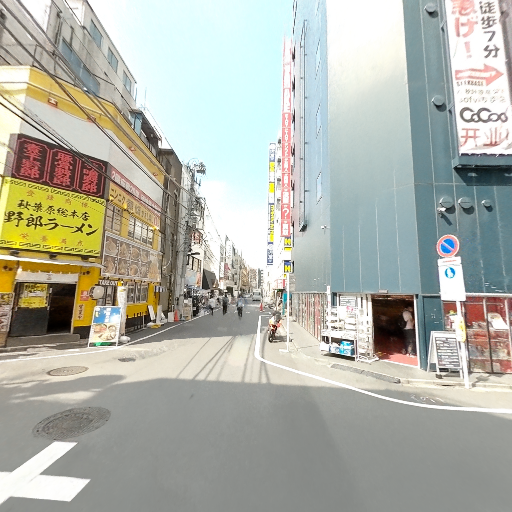} \\
            \includegraphics[width=\itemwidth]{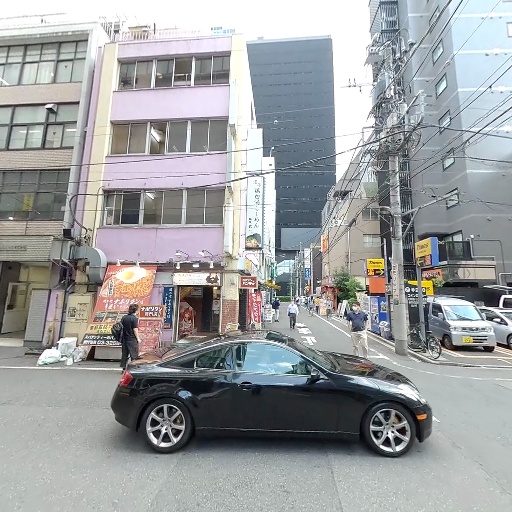} &
            \includegraphics[width=\itemwidth]{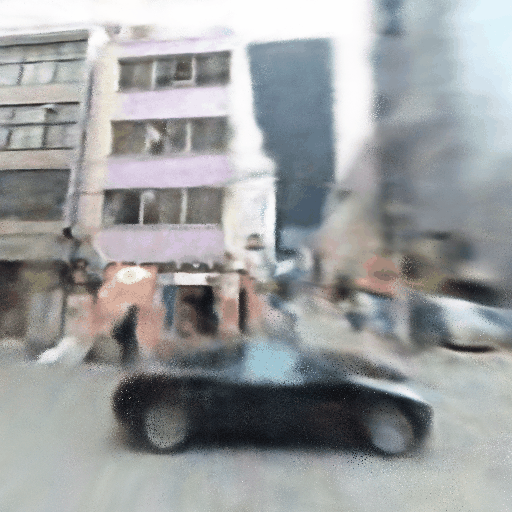} &
            \includegraphics[width=\itemwidth]{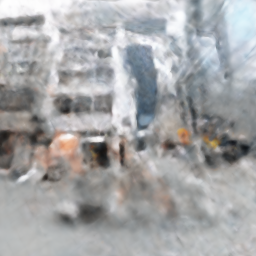} &
            \includegraphics[width=\itemwidth]{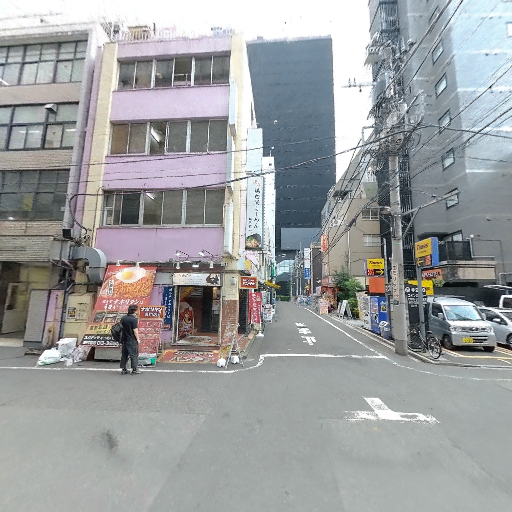} \\\vspace{0.2em}
            Original image & $\mathrm{D^2}$NeRF~\cite{d2nerf} & RoDynRF~\cite{liu2023robust} & Entity-NeRF \\
        \\
    \end{tabular}\vspace{-1.5em}
  \caption{\textbf{Qualitative comparison with dynamic NeRF methods ($\mathbf{D^2}$NeRF~\cite{d2nerf} and  RoDynRF~\cite{liu2023robust}) on MovieMap Dataset.}}
  \label{fig:vis_dynamic_nerf}
\end{figure}

\noindent \textbf{MovieMap Dataset}:
\href{https://moviemap.jp/}{Movie Map}~\cite{moviemap} offers an immersive interface for walking through cities in Japan using 360\textdegree~videos. Movie Map currently offers the exploration of eight cities. From them, 360\textdegree~videos of the Akihabara scene are used to create the dataset. Akihabara is characterized by an especially high number of pedestrians, vehicles, bicycles, and various moving objects, even by global standards. Additionally, the background comprises buildings of various shapes and colorful outdoor advertisements, embodying all the typical urban characteristics that make learning with Neural Radiance Fields (NeRF) challenging.

Our MovieMap Dataset comprises three different subsets, containing 12, 15, and 51 images each, sampled from the original 360\textdegree~video in Movie Map. Manual annotation of moving objects was performed on each 360\textdegree~image. As illustrated in \Fref{fig:vis_our_data}, stationary cars and pedestrians are not labeled as distractors. For evaluation, we generated images without distractors by using Nerfacto~\cite{nerfstudio} with annotated distractor labels to exclude moving objects from the training. This process allows us to render images from the same camera viewpoints to create background-only images.

In this study, we extract 14 perspective images from each 360\textdegree~image, utilizing their degree of field (DoF) as input parameters for NeRF training. A full description of MovieMap Dataset is in the supplementary material.

\subsection{Evaluation with dynamic NeRF models on MovieMap Dataset}
To show urban environments' unsuitability for NeRF models encoding static and dynamic data, we tested $\mathrm{D^2}$NeRF~\cite{d2nerf} and RoDynRF~\cite{liu2023robust} as dynamic NeRF methods. For RoDynRF, designed solely for monocular video, we converted 360\textdegree~images into 90\textdegree~perspective images from the image center.

\Fref{fig:vis_dynamic_nerf} demonstrates the qualitative results using $\mathrm{D^2}$NeRF and RoDynRF in urban settings with motion. Both models struggle with moving objects and static background reconstruction, due to their limitations in handling excessive movement, complex motion, and scale variations. This makes developing dynamic NeRF models for urban scenes problematic. Subsequent experiments compare our work with RobustNeRF, which ignores moving objects.

\begin{figure}[t]
    \centering
    \setlength{\tabcolsep}{0.02cm}
    \setlength{\itemwidth}{4.1cm}
    \renewcommand{\arraystretch}{0.5}
    \hspace*{-\tabcolsep}\small\begin{tabular}{cc}
            \includegraphics[width=\itemwidth]{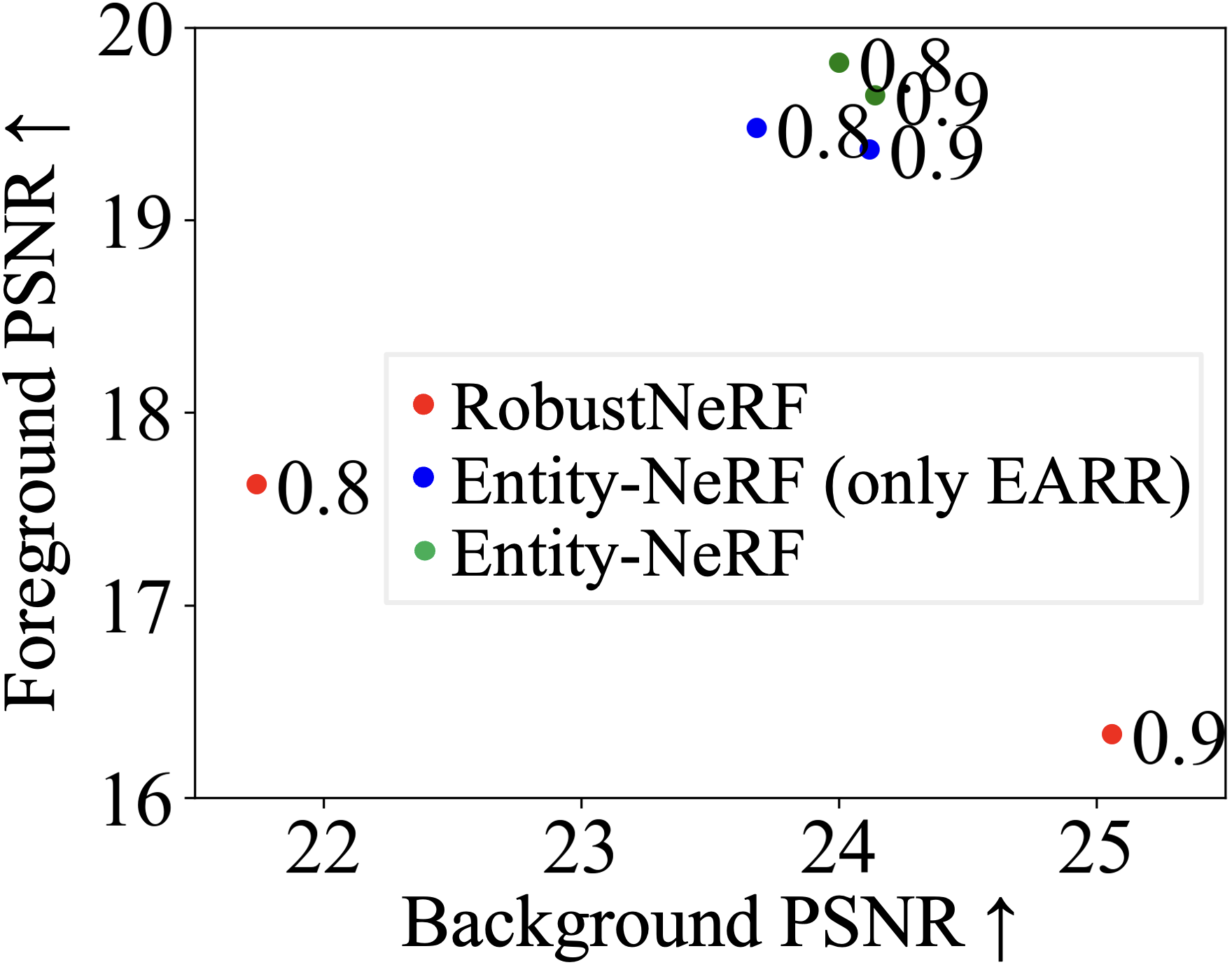} &
            \includegraphics[width=\itemwidth]{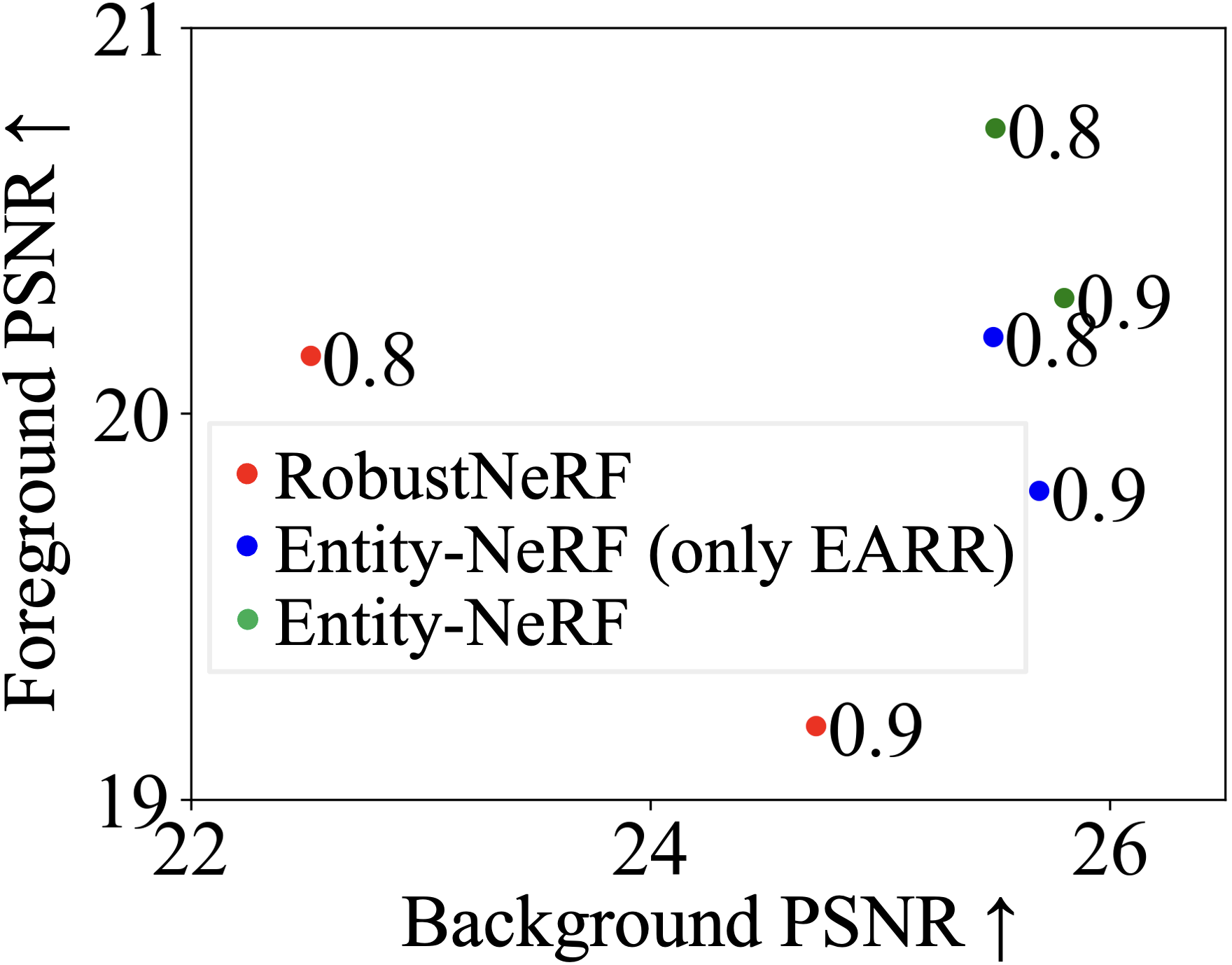} \\\vspace{0.2em}
            Nerfacto~\cite{nerfstudio} &
            Mip-NeRF 360~\cite{barron2022mipnerf360} \\
        \\
    \end{tabular}\vspace{-1.5em}
  \caption{\textbf{The trade-off between the foreground/background PSNR.} The values in the figure indicate the threshold of inliers.}
  \label{fig:trade_off}
\end{figure}

\subsection{Evaluation on MovieMap Dataset}
\noindent\textbf{Qualitative comparison:}
A comparison of RobustNeRF~\cite{robustnerf} and our proposed method using Nerfacto~\cite{nerfstudio} and Mip-NeRF 360~\cite{barron2022mipnerf360} on urban scenes is shown in \Tref{table:experiment_realdata}. The mean-squared error (MSE) of incorporating all static backgrounds and moving objects into training enhances the PSNR of the backgrounds, which make up a larger percentage, leading to an increased overall PSNR. Therefore, it is crucial to evaluate foreground and background PSNR separately. Our proposed method achieved a background PSNR close to the mean-squared error while exceeding existing methods in foreground PSNR. The comparison with RobustNeRF showed consistent improvements.\\

\begin{figure}[t]
    \centering
    \setlength{\tabcolsep}{0.1cm}
    \setlength{\itemwidth}{4.1cm}
    \renewcommand{\arraystretch}{0.5}
    \hspace*{-\tabcolsep}\small\begin{tabular}{cc}
            \includegraphics[width=\itemwidth]{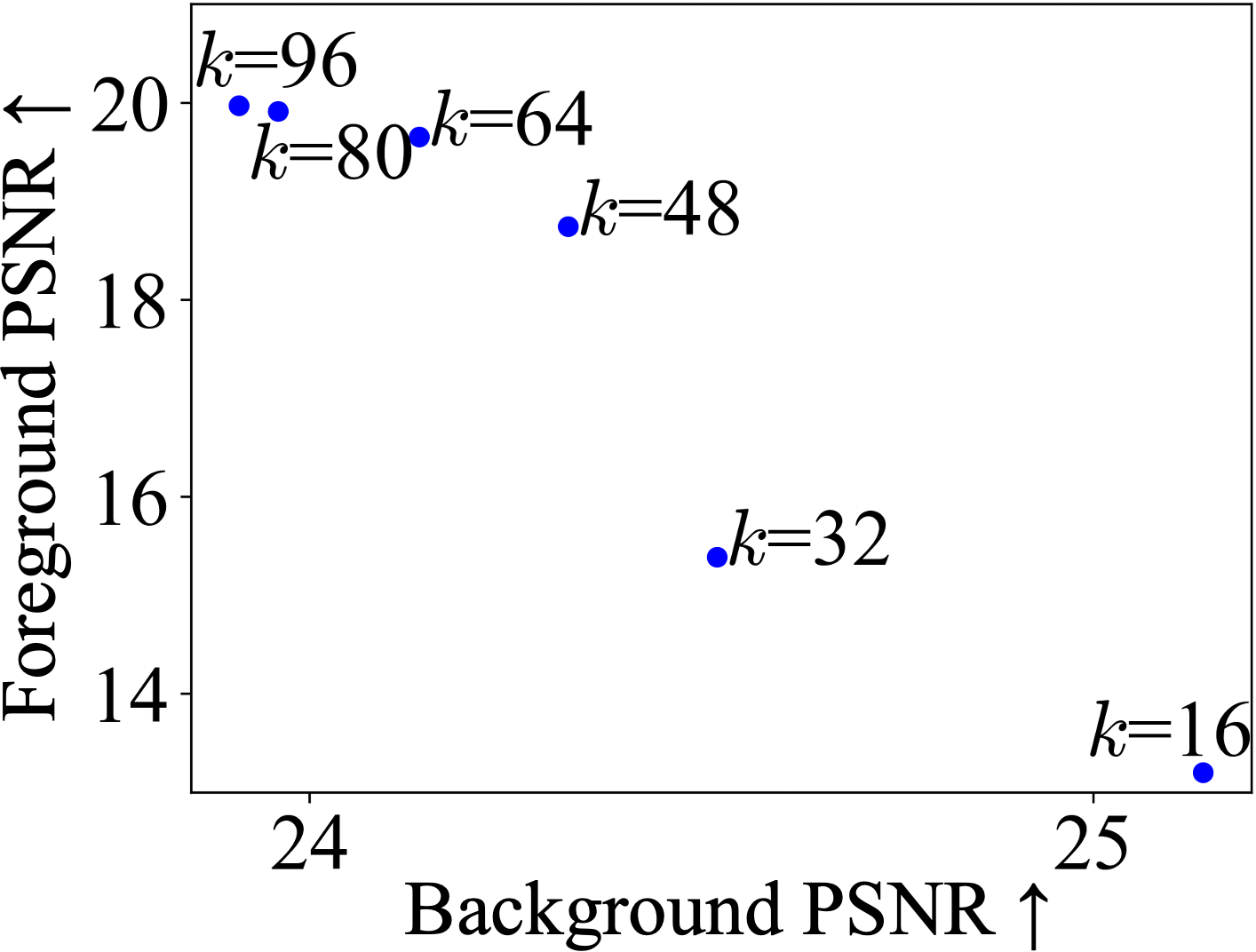} &
            \includegraphics[width=\itemwidth]{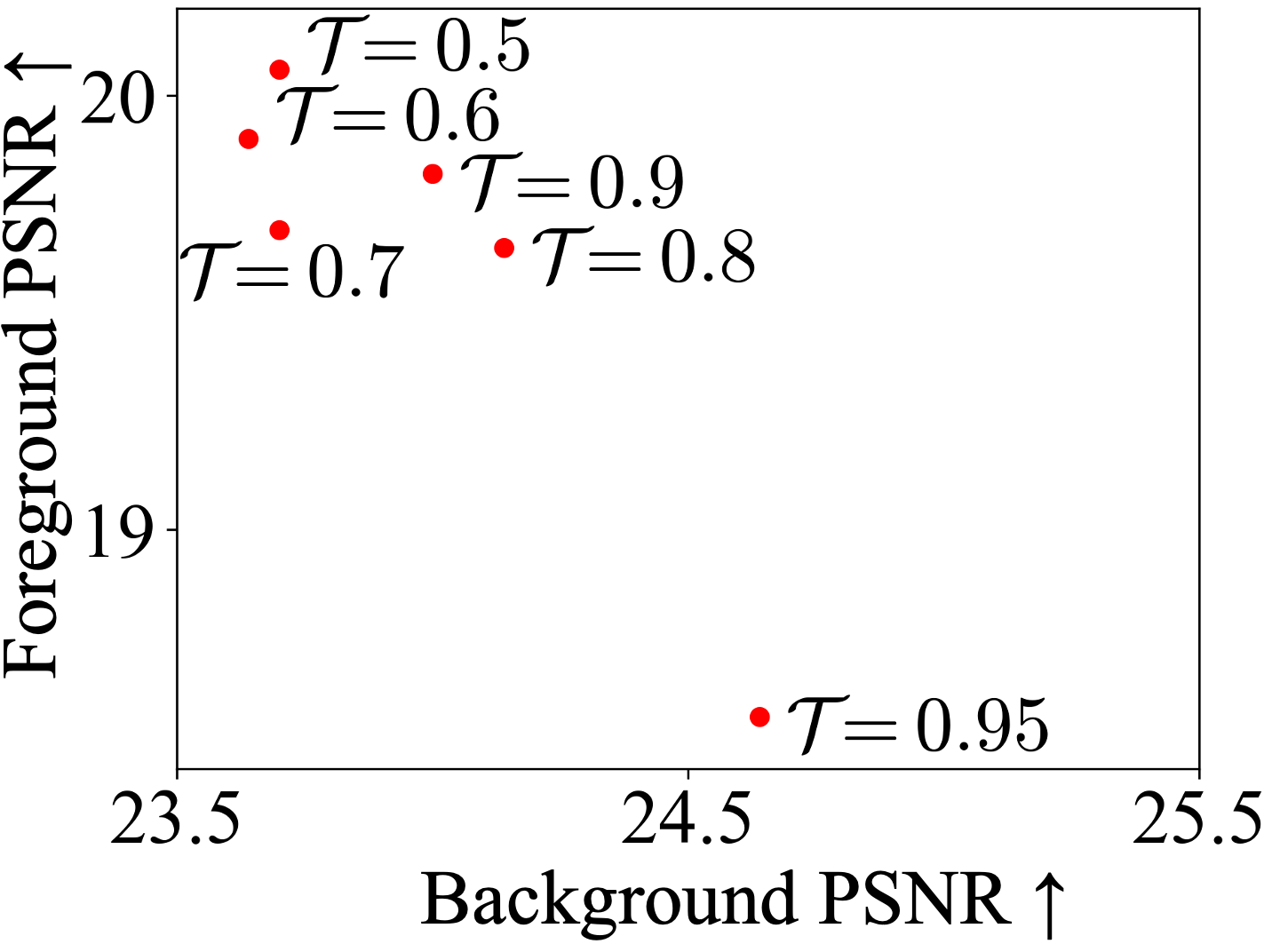} \\\vspace{0.2em}
    \end{tabular}\vspace{-2.0em}
  \caption{\textbf{Sensitivity to patch size ($\bm{k}$) and threshold ($\bm{\mathcal{T}}$).}}
  \label{fig:sensitivity}
\end{figure}

\begin{figure}[t]
    \centering
    \setlength{\tabcolsep}{0.1cm}
    \setlength{\itemwidth}{4.1cm}
    \renewcommand{\arraystretch}{0.5}
    \hspace*{-\tabcolsep}\small\begin{tabular}{cc}
            \includegraphics[width=\itemwidth]{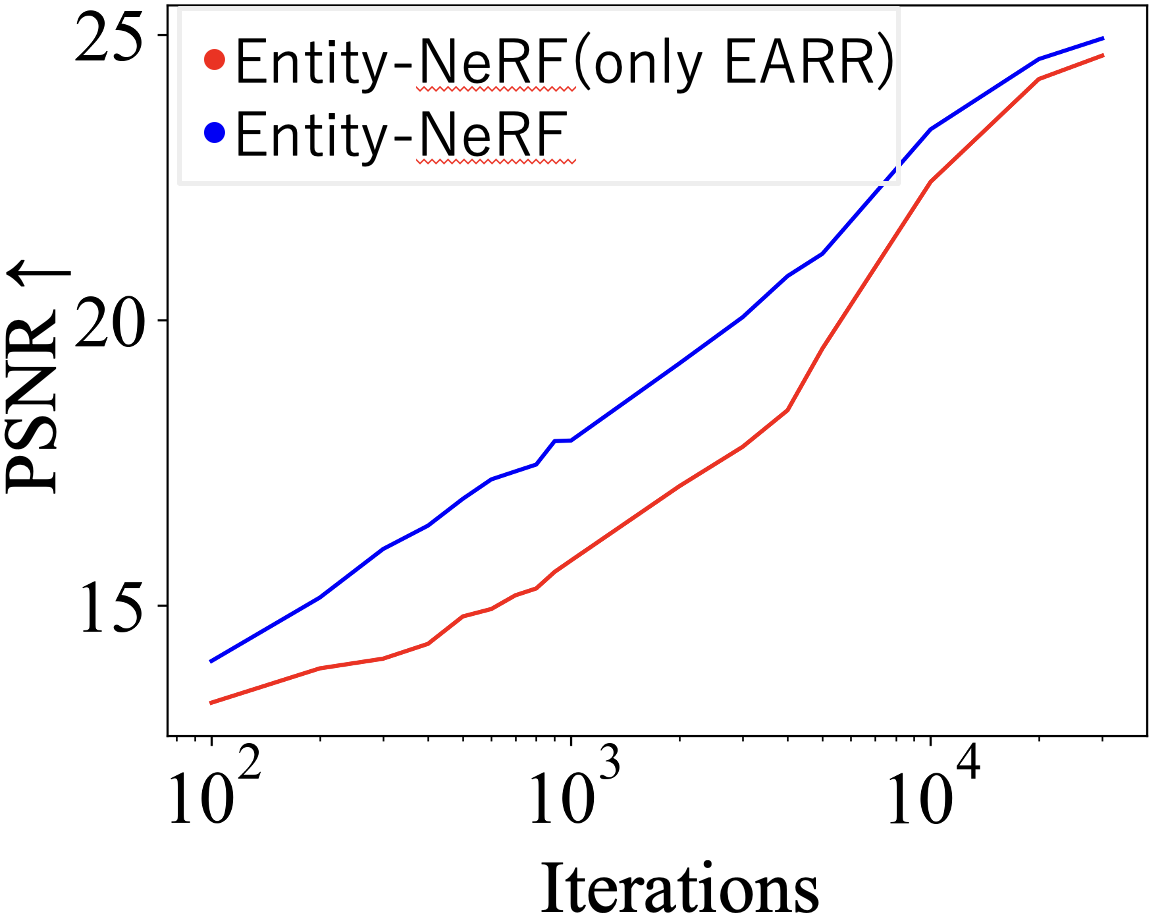} &
            \includegraphics[width=\itemwidth]{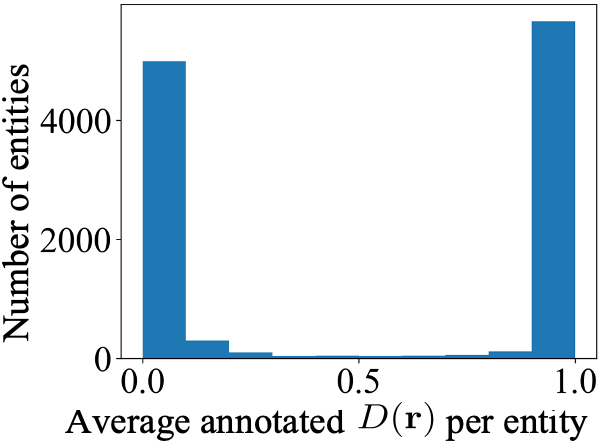} \\\vspace{0.2em}
            \begin{minipage}{0.2\textwidth}\caption{\label{fig:iteration}\textbf{Difference in the training curves.}}\end{minipage} & \begin{minipage}{0.2\textwidth}\vspace{-0.2em}\caption{\label{fig:mean_distractiveness}\textbf{Histogram of the average $\bm{D(}\mathbf{r}\bm{)}$ per entity.}}\end{minipage} \\
            \\
    \end{tabular}\vspace{-1.5em}
\end{figure}

\noindent\textbf{Sensitivity of hyperparameters:} RobustNeRF and our proposed method present a trade-off between the foreground PSNR and background PSNR, influenced by a hyperparameter that determines the inlier to the residual ratio (denoted by $\mathcal{T}$ in our EARR). This trade-off is shown in the \Fref{fig:trade_off}. Raising the inlier ratio improves background PSNR, but risks including moving objects in the learning, decreasing foreground PSNR. Similarly, lowering the inlier ratio worsens background PSNR, but removes many moving objects, boosting foreground PSNR. Entity-NeRF shows consistent improvements in foreground PSNR for Nerfacto and in background PSNR for Mip-NeRF 360. In addition, while RobustNeRF is biased toward improving one of the metrics, Entity-NeRF achieves more balanced results by increasing both metrics.

In addition, we performed a detailed sensitivity analysis on hyperparameters (\ie, patch size $k$ and threshold $\mathcal{T}$ in EARR). As shown in~\cref{fig:sensitivity}, increasing $k$ improved the foreground PSNR with only a minor background PSNR impact. The choice of $\mathcal{T}$ proved less sensitive than $k$, provided it remains below the typical inlier ratio in urban scenes (\eg, 90.4\% in MovieMap dataset).\\

\noindent\textbf{Effects of stationary entity classification:} We conducted an analysis comparing the training curves of Entity-NeRF with and without stationary entity classification. As shown in \cref{fig:iteration}, stationary entity classification not only significantly boosts training efficiency but also enhances final PSNR.\\

\noindent\textbf{Validity of entity segmentation:} To confirm the stable performance of our entity segmentation across various images, we calculated the average of annotated labels for each segmented entity using all images in the MovieMap dataset. Then, we constructed a histogram representing these average values for all entities. As depicted in~\cref{fig:mean_distractiveness}, the distribution is noticeably skewed towards either 0 or 1, which indicates that entities are clearly segmented into moving ($=0$) or static ($=1$) entities.\\

\begin{figure*}[t]
    \centering
    \setlength{\tabcolsep}{0.02cm}
    \setlength{\itemwidth}{2.4cm}
    \renewcommand{\arraystretch}{0.5}
    \hspace*{-\tabcolsep}\small\begin{tabular}{ccccccc}
            \includegraphics[width=\itemwidth]{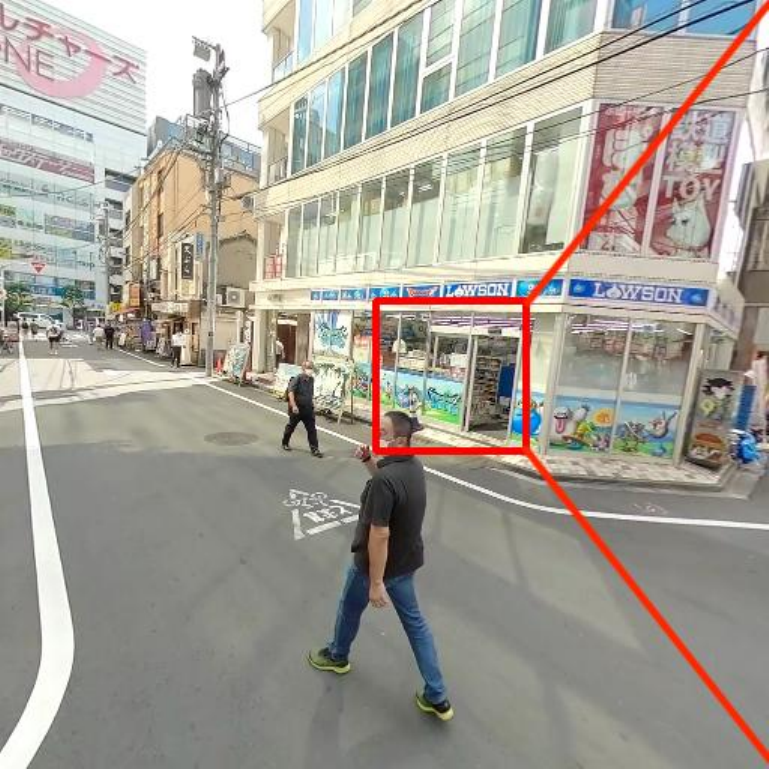} &
            \fboxsep=0pt\fcolorbox{red}{white}{\includegraphics[width=\itemwidth]{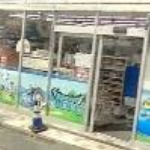}} &
            \includegraphics[width=\itemwidth]{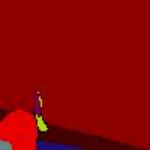} &
            \fboxsep=0pt\fbox{\includegraphics[width=\itemwidth]{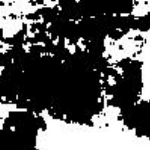}} &
            \fboxsep=0pt\fcolorbox{red}{white}{\includegraphics[width=\itemwidth]{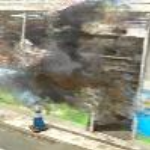}} &
            \fboxsep=0pt\fbox{\includegraphics[width=\itemwidth]{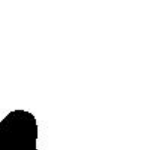}} &
            \fboxsep=0pt\fcolorbox{red}{white}{\includegraphics[width=\itemwidth]{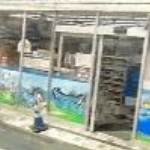}} \\
            \includegraphics[width=\itemwidth]{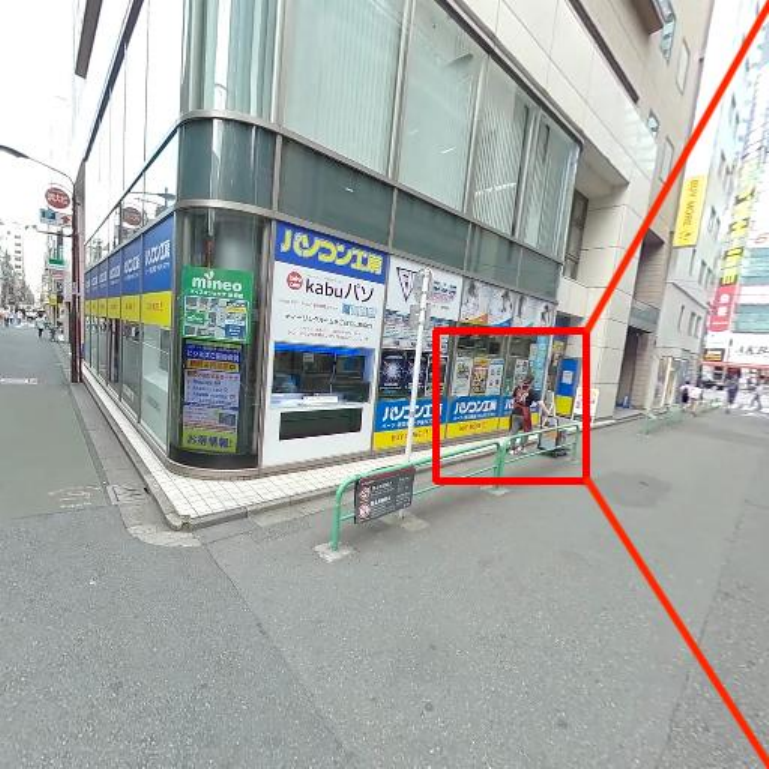} &
            \fboxsep=0pt\fcolorbox{red}{white}{\includegraphics[width=\itemwidth]{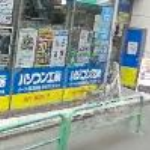}} &
            \includegraphics[width=\itemwidth]{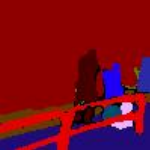} &
            \fboxsep=0pt\fbox{\includegraphics[width=\itemwidth]{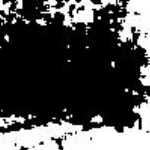}} &
            \fboxsep=0pt\fcolorbox{red}{white}{\includegraphics[width=\itemwidth]{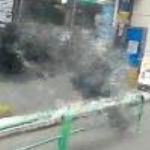}} &
            \fboxsep=0pt\fbox{\includegraphics[width=\itemwidth]{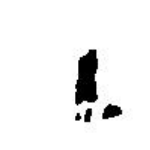}} &
            \fboxsep=0pt\fcolorbox{red}{white}{\includegraphics[width=\itemwidth]{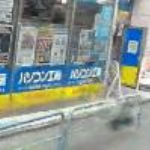}} \\
            \includegraphics[width=\itemwidth]{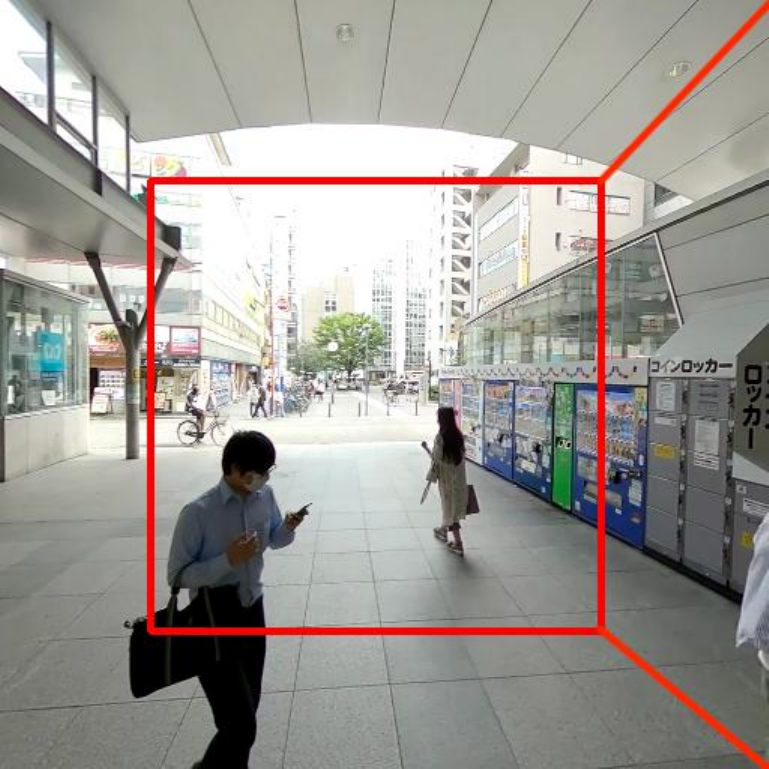} &
            \fboxsep=0pt\fcolorbox{red}{white}{\includegraphics[width=\itemwidth]{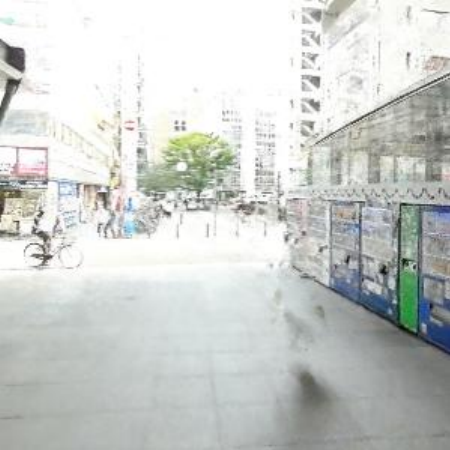}} &
            \includegraphics[width=\itemwidth]{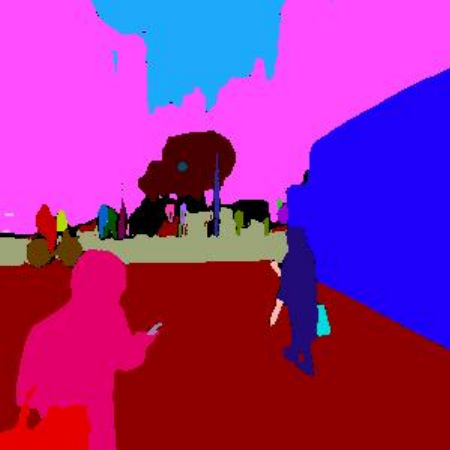} &
            \fboxsep=0pt\fbox{\includegraphics[width=\itemwidth]{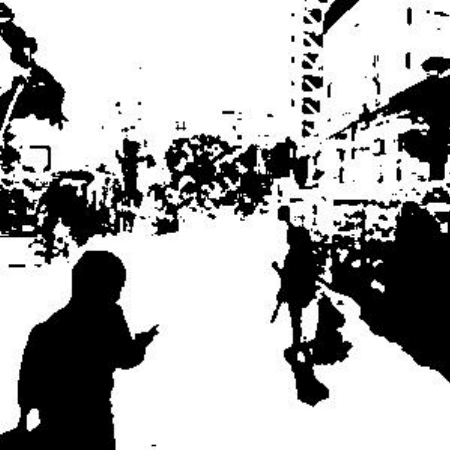}} &
            \fboxsep=0pt\fcolorbox{red}{white}{\includegraphics[width=\itemwidth]{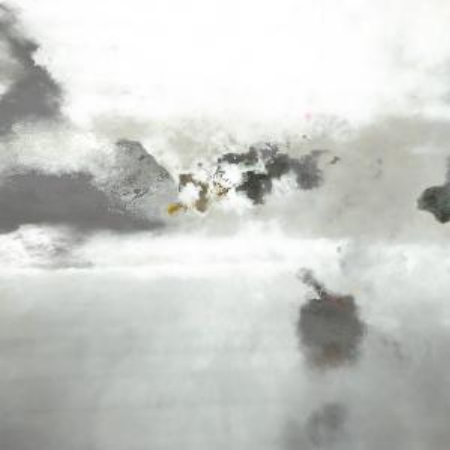}} &
            \fboxsep=0pt\fbox{\includegraphics[width=\itemwidth]{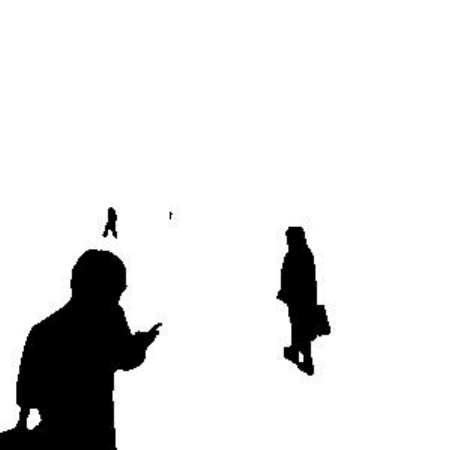}} &
            \fboxsep=0pt\fcolorbox{red}{white}{\includegraphics[width=\itemwidth]{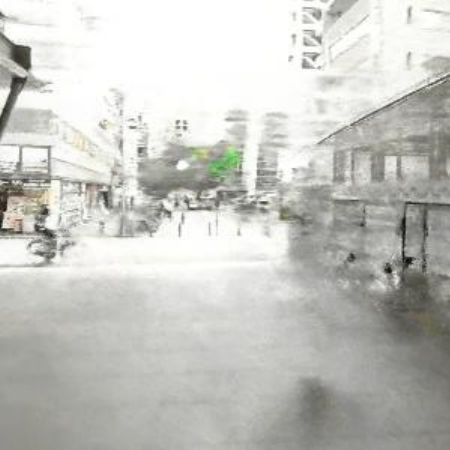}} \\
            \includegraphics[width=\itemwidth]{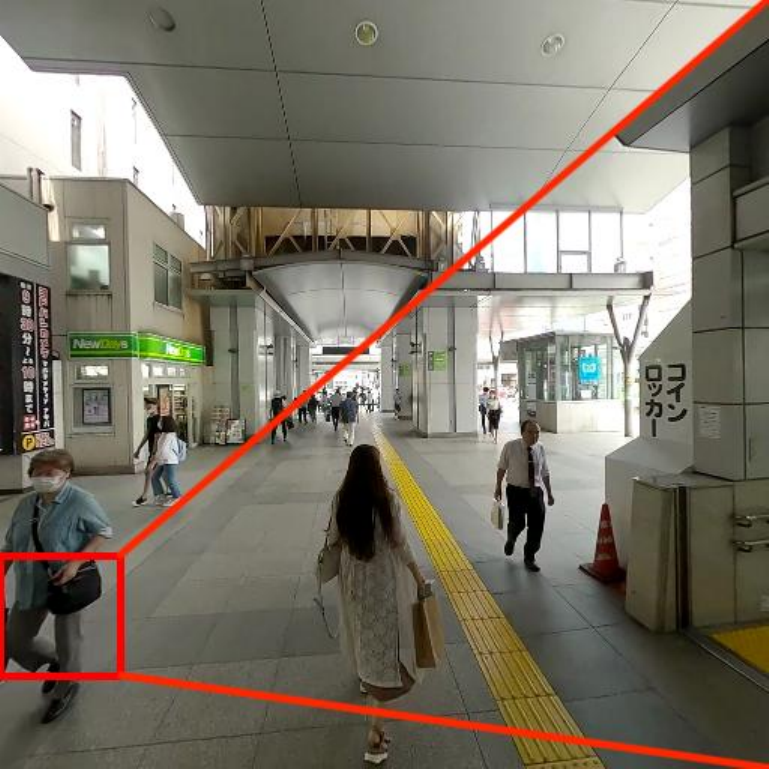} &
            \fboxsep=0pt\fcolorbox{red}{white}{\includegraphics[width=\itemwidth]{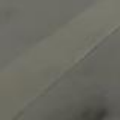}} &
            \includegraphics[width=\itemwidth]{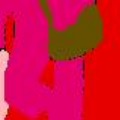} &
            \fboxsep=0pt\fbox{\includegraphics[width=\itemwidth]{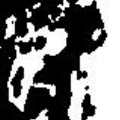}} &
            \fboxsep=0pt\fcolorbox{red}{white}{\includegraphics[width=\itemwidth]{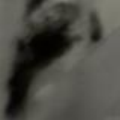}} &
            \fboxsep=0pt\fbox{\includegraphics[width=\itemwidth]{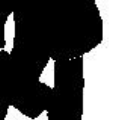}} &
            \fboxsep=0pt\fcolorbox{red}{white}{\includegraphics[width=\itemwidth]{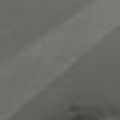}} \\
            \vspace{0.2em}
            Original image &
            Rendered &
            Entity Seg.~\cite{qilu2023high} &
            RobustNeRF~\cite{robustnerf} &
            RobustNeRF~\cite{robustnerf} &
            Entity-NeRF &
            Entity-NeRF \\
            & ground-truth & & $D(\mathbf{r})$ & & $D(\mathbf{r})$ & \\
        \\
    \end{tabular}\vspace{-1.5em}
  \caption{\textbf{Qualitative comparison including Entity Seg.~\cite{qilu2023high} on MovieMap Dataset.} $D(\mathbf{r})$ is calculated at the end of the training.}
  \label{fig:exp_real_data}
\end{figure*}

\noindent\textbf{Qualitative comparison:}
A qualitative comparison with RobustNeRF~\cite{robustnerf} is shown in \Fref{fig:exp_real_data}. Our Entity-NeRF successfully reconstructed complex building walls that RobustNeRF mistakenly removed (\Fref{fig:exp_real_data}-top three items). It also effectively removed moving objects that RobustNeRF failed to remove (\Fref{fig:exp_real_data}-bottom two items). Thus, Entity-NeRF is clearly superior to RobustNeRF in both removing moving objects and reconstructing static backgrounds.

\begin{figure}[t]
    \centering
    \setlength{\tabcolsep}{0.02cm}
    \setlength{\itemwidth}{2.7cm}
    \renewcommand{\arraystretch}{0.5}
    \hspace*{-\tabcolsep}\begin{tabular}{ccc}
            \includegraphics[width=\itemwidth]{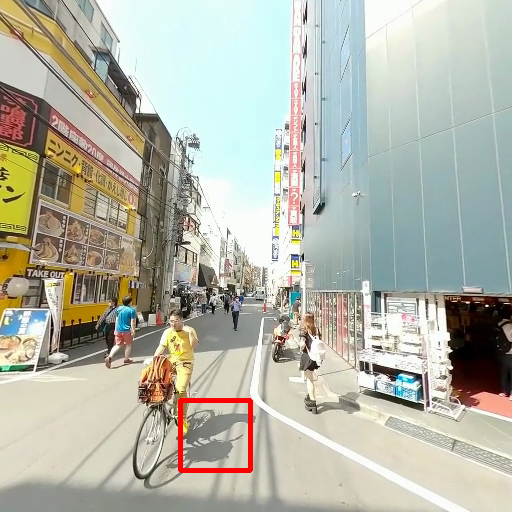} &
            \includegraphics[width=\itemwidth]{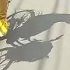} &
            \includegraphics[width=\itemwidth]{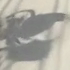} \\
            Original image & Original image & Entity-NeRF\\
    \end{tabular}
  \vspace{-0.25cm}\caption{\textbf{Limitations.} Entity-NeRF cannot handle shadows.}
  \label{fig:shadow}
\end{figure}

\section{Conclusion}
\label{sec:conclusion}
We address the problem of identifying and removing multiple moving objects of various categories and scales to build a NeRF for dynamic urban scenes. To solve this problem, we introduce Entity-wise Average of Residual Ranks designed to identify moving objects using entity-wise statistics and the stationary entity classification with thing/stuff segmentation to remove complex backgrounds in the early stages of NeRF training. Our evaluation using an urban scene dataset, where existing methods fail to model scene dynamics or remove moving objects, shows quantitatively and qualitatively that the proposed method works very well.\\

\noindent\textbf{Limitations}:
While Entity-NeRF demonstrates outstanding performance in urban environments, it is subject to a few limitations. Firstly, if a large moving object dominates the image and thereby obscures the background from another perspective, there might be difficulty in successfully reconstructing the background hidden by the moving object. This issue, however, could potentially be mitigated by integrating existing inpainting techniques. 

Moreover, as shown in \Fref{fig:shadow}, since shadows are not explicitly managed in the current framework, shadows cast by moving objects might be inadvertently incorporated into the training process. This issue may be resolved by using segmentation that includes shadows or by removing shadows in post-processing.

\section*{Acknowledgments}
\label{sec:acknowledgments}
This research was partly supported by JST Mirai JPMJMI21H1, JSPS KAKENHI 21H03460, and CISTI SIP.

\clearpage

{
    \small
    \bibliographystyle{ieee_fullname}
    \bibliography{egbib}
}

\input{sec/X_suppl}

\end{document}

%% file: sec/X_suppl.tex
\clearpage
\setcounter{page}{1}
\maketitlesupplementary

\section{MovieMap Dataset Details}
\label{sec:dataset_details}
\begin{figure}[t]
    \centering
    \setlength{\tabcolsep}{0.02cm}
    \setlength{\itemwidth}{2.0cm}
    \renewcommand{\arraystretch}{0.5}
    \hspace*{-\tabcolsep}\small\begin{tabular}{cccc}
            \multicolumn{4}{c}{\includegraphics[width=8.1cm]{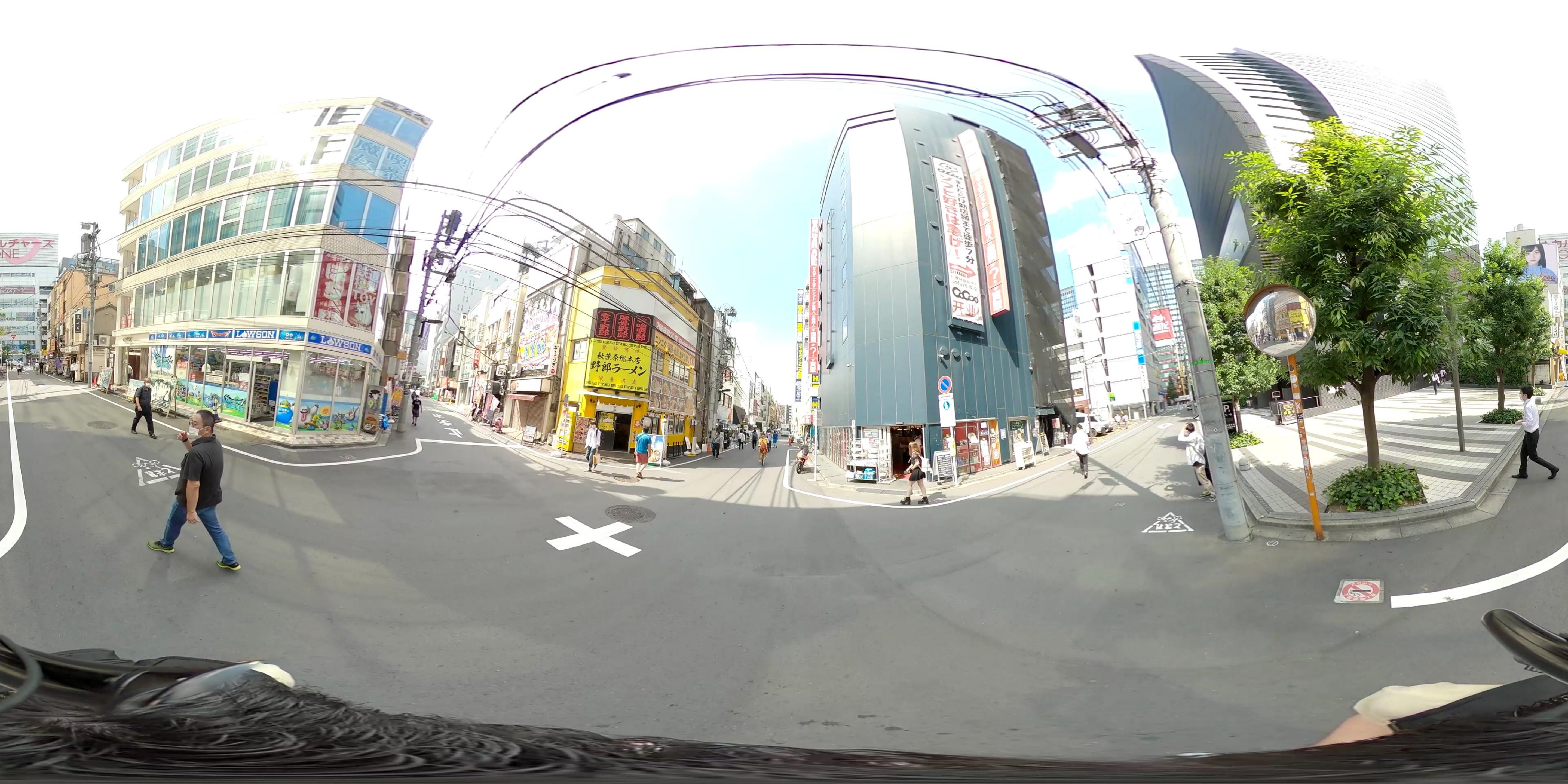}}\\
            \multicolumn{4}{c}{360\textdegree~image}\\\\
            \includegraphics[width=\itemwidth]{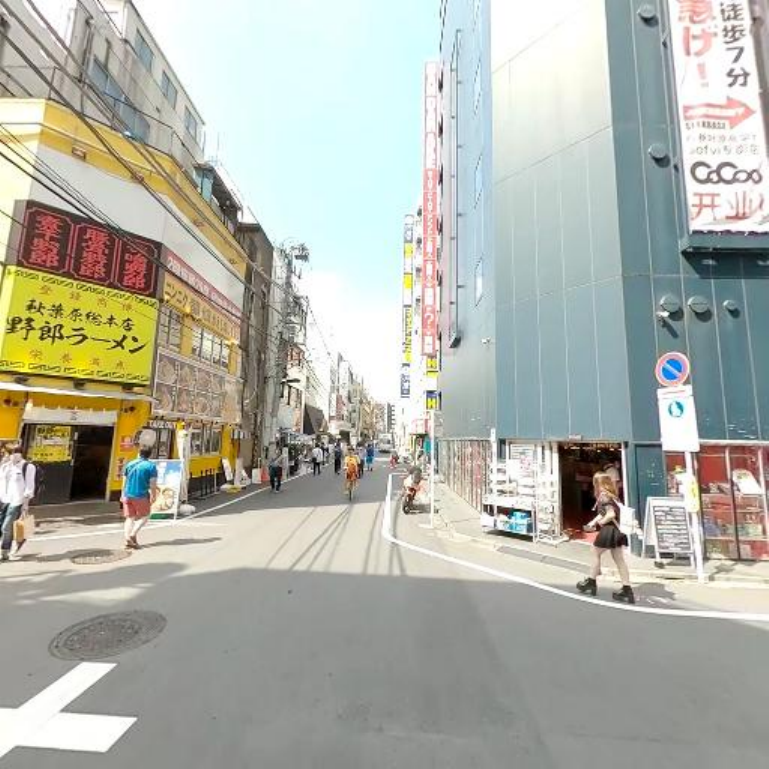} &
            \includegraphics[width=\itemwidth]{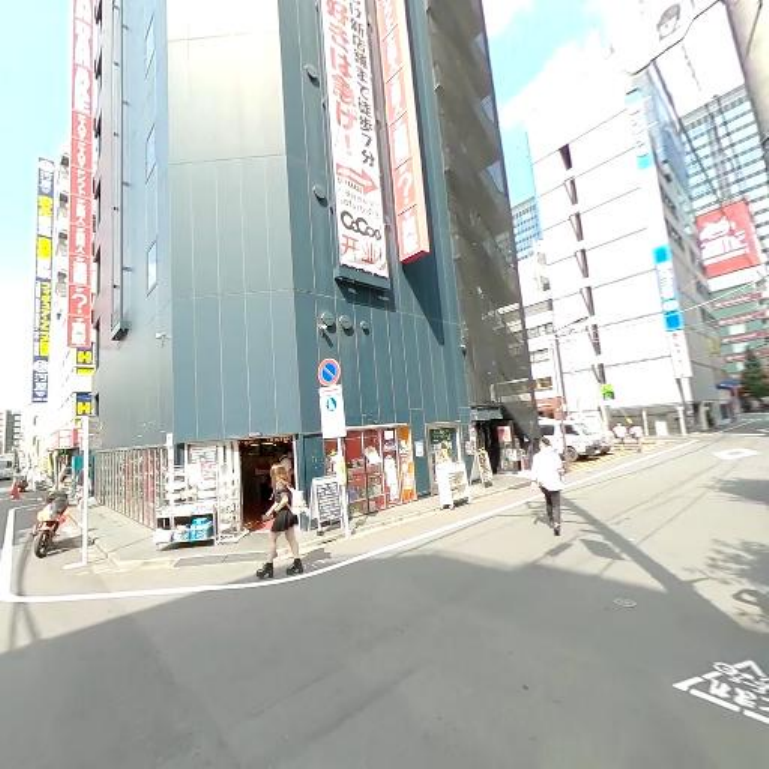} &
            \includegraphics[width=\itemwidth]{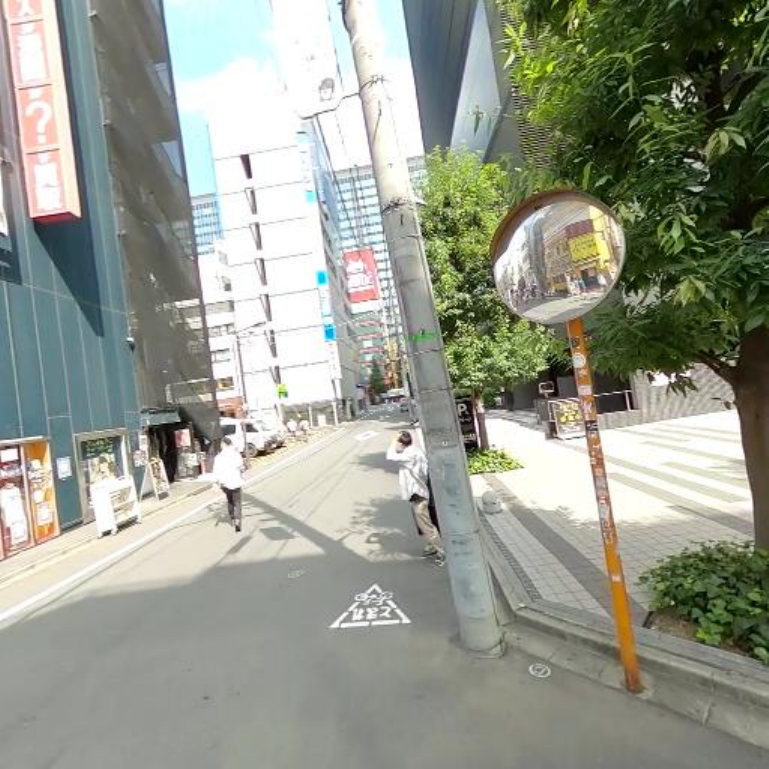} &
            \includegraphics[width=\itemwidth]{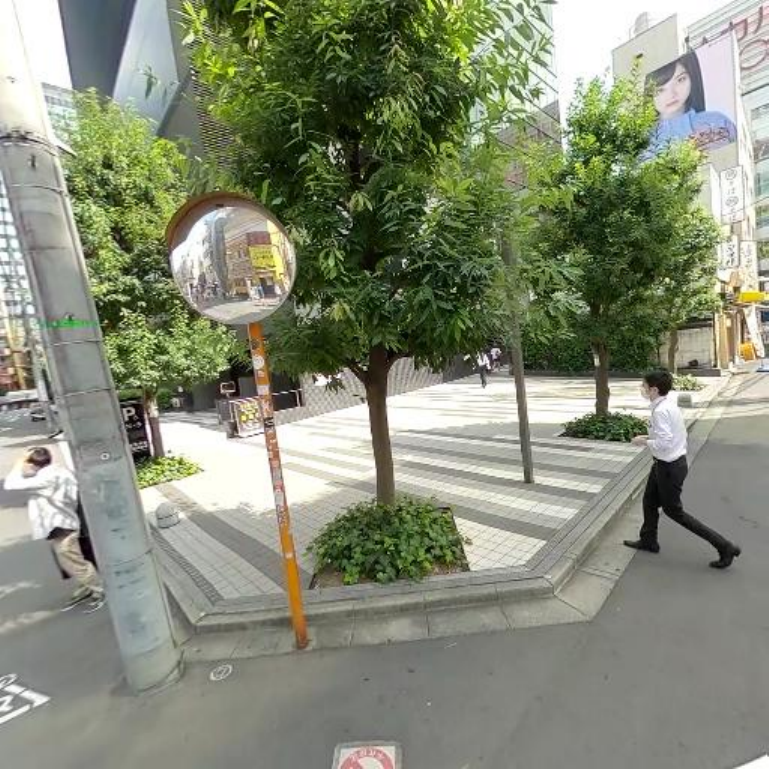} \\
            \includegraphics[width=\itemwidth]{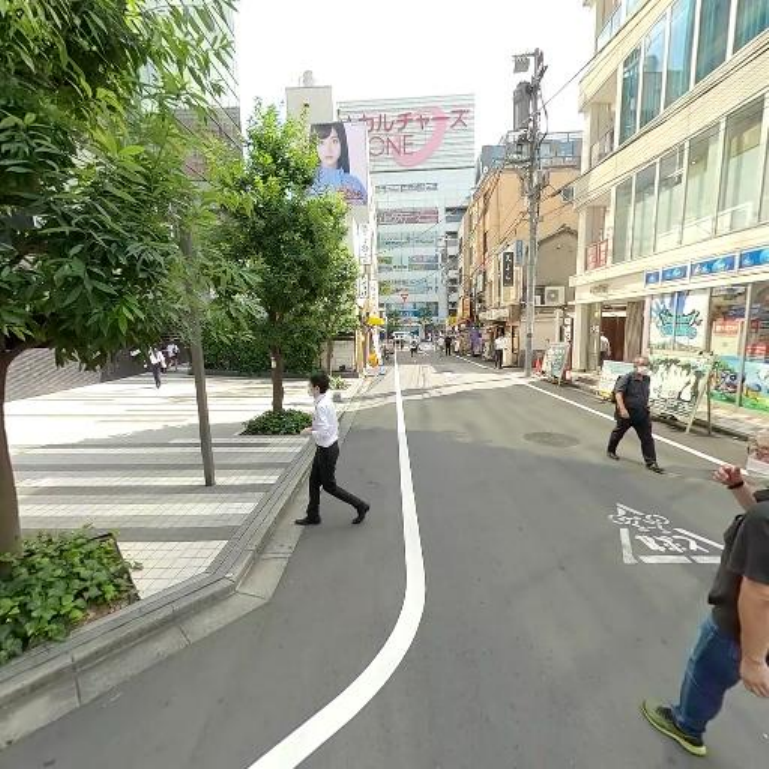} &
            \includegraphics[width=\itemwidth]{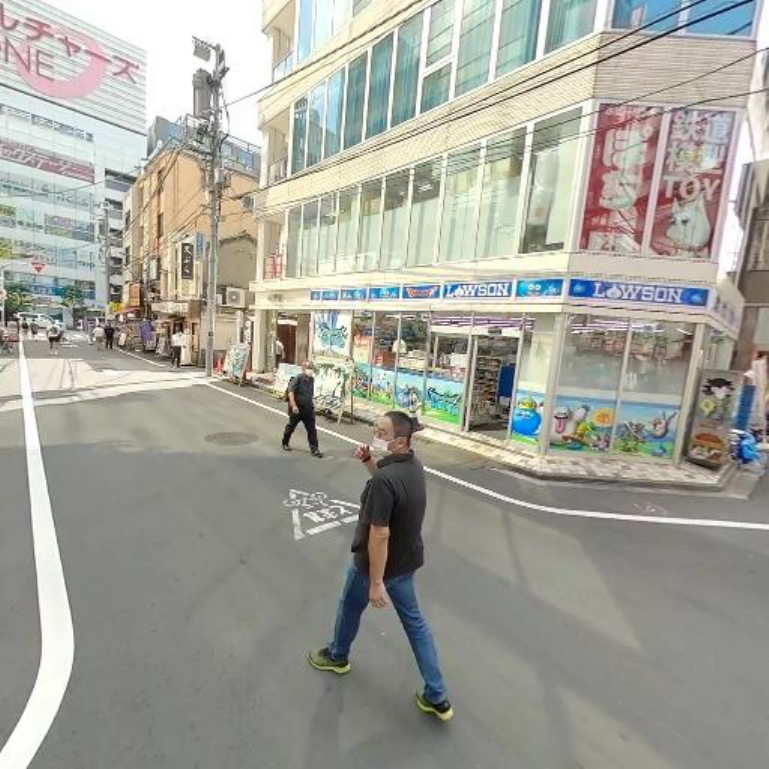} &
            \includegraphics[width=\itemwidth]{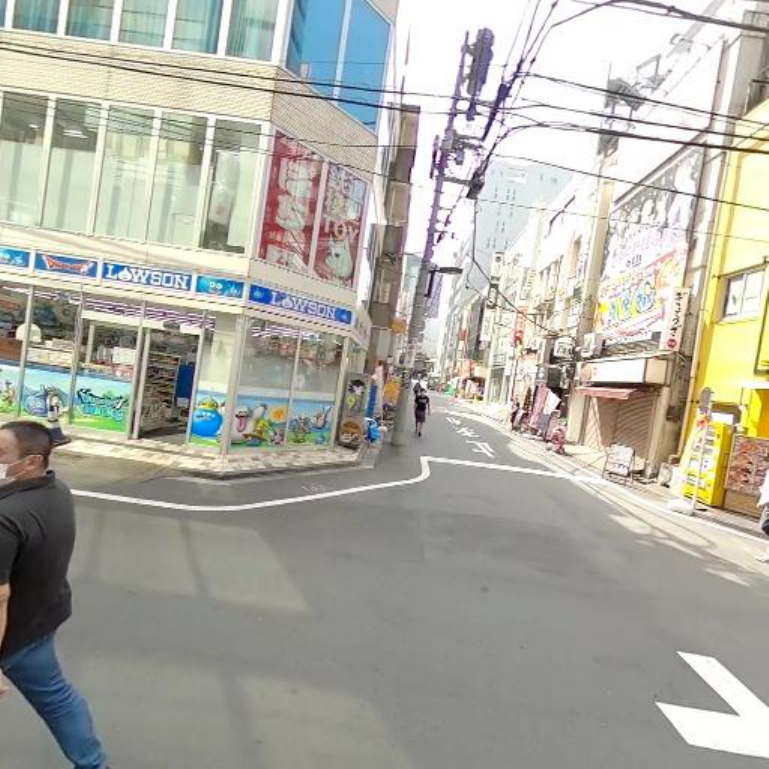} &
            \includegraphics[width=\itemwidth]{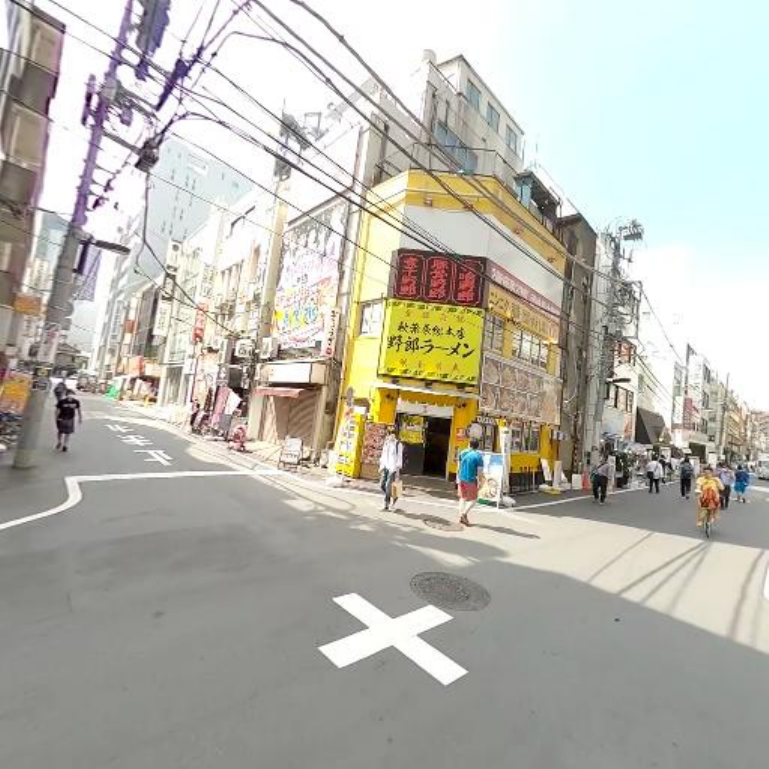} \\
            \includegraphics[width=\itemwidth]{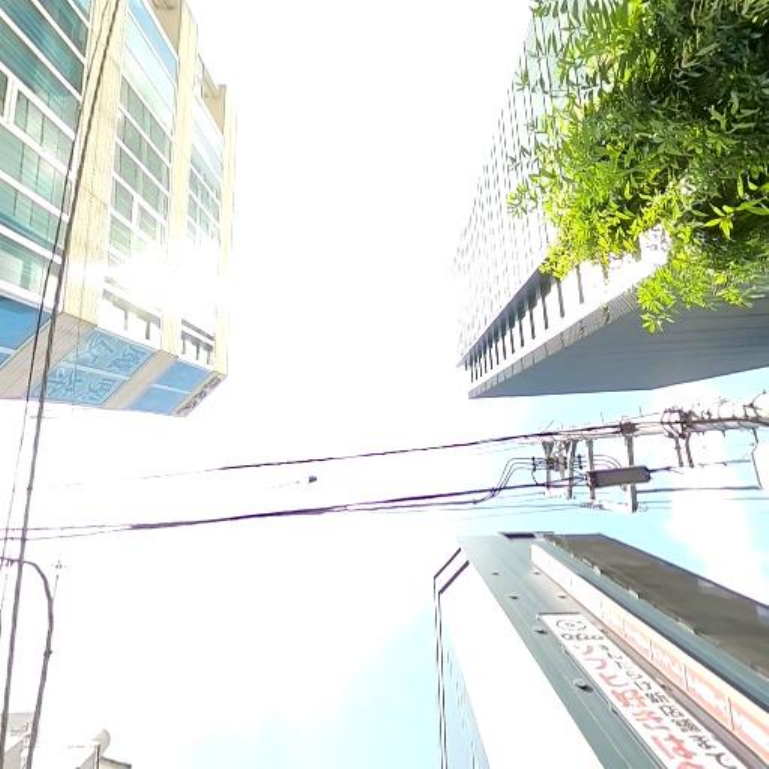} &
            \includegraphics[width=\itemwidth]{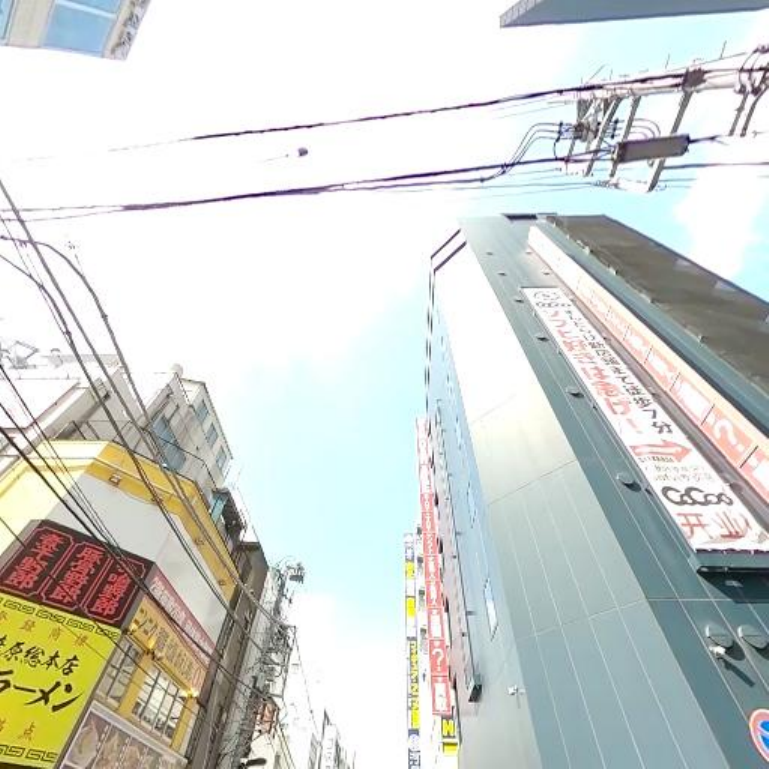} &
            \includegraphics[width=\itemwidth]{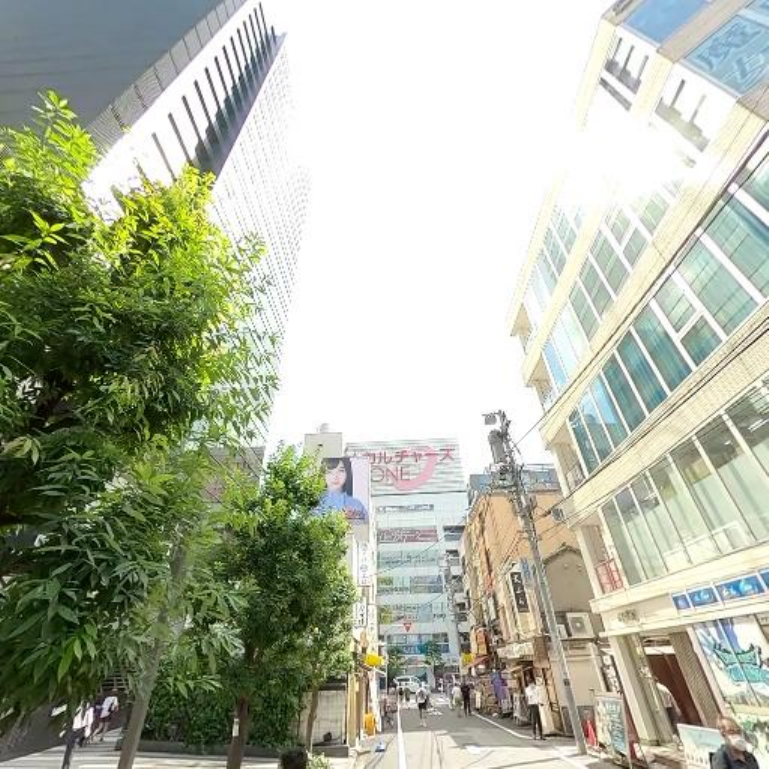} &
            \includegraphics[width=\itemwidth]{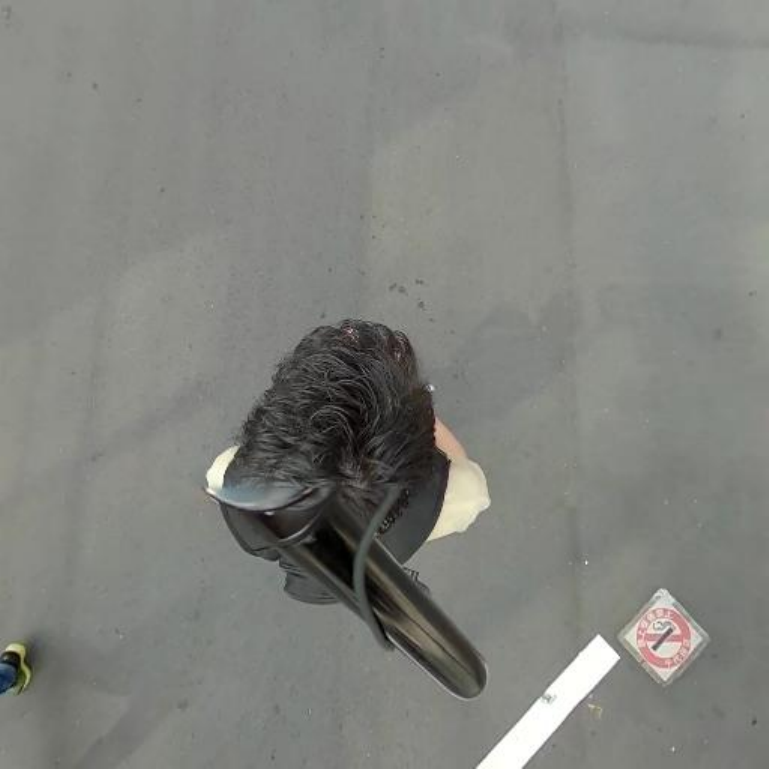} \\
            & \includegraphics[width=\itemwidth]{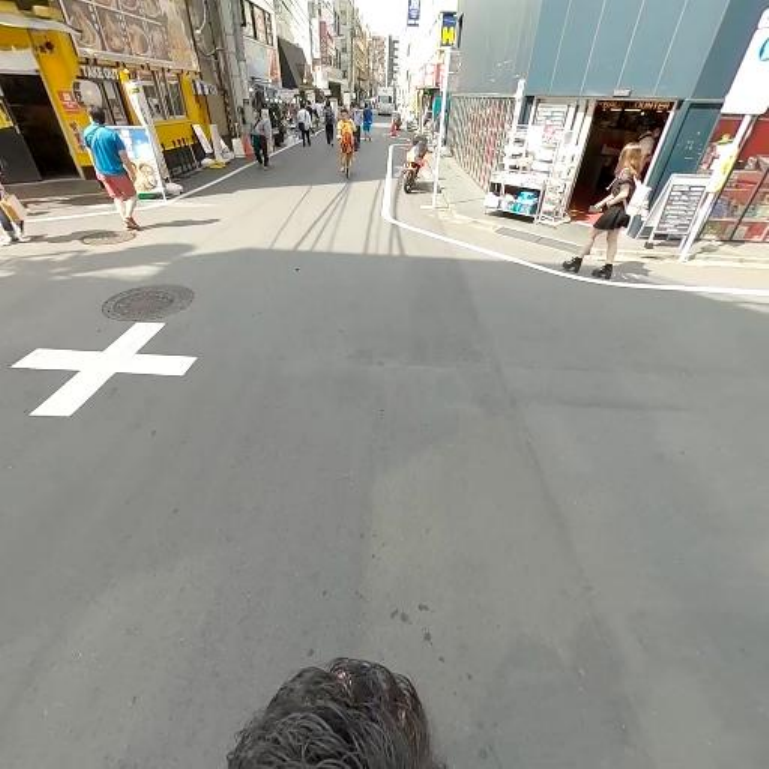} &
            \includegraphics[width=\itemwidth]{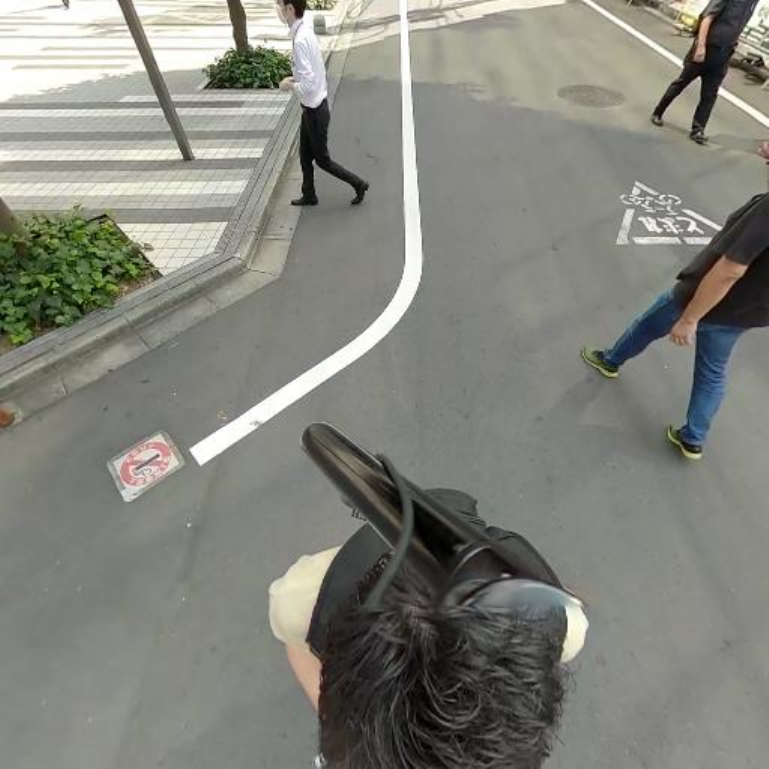}
            \\
            \multicolumn{4}{c}{Perspective projection images}\\\vspace{0.2em}
        \\
    \end{tabular}\vspace{-1.5em}
  \caption{\textbf{Perspective projection images extracted from a single 360\textdegree~image.}}
  \label{fig:cut_out_pers}
\end{figure}

\begin{figure}[t]
    \centering
    \setlength{\tabcolsep}{0.02cm}
    \setlength{\itemwidth}{2.7cm}
    \renewcommand{\arraystretch}{0.5}
    \hspace*{-\tabcolsep}\begin{tabular}{ccc}
            \includegraphics[width=\itemwidth]{figures/supp/split/frame_00012.pdf} &
            \includegraphics[width=\itemwidth]{figures/supp/split/frame_00013.pdf} &
            \includegraphics[width=\itemwidth]{figures/supp/split/frame_00014.pdf} \\
            \fboxsep=0pt\fbox{\includegraphics[width=\itemwidth]{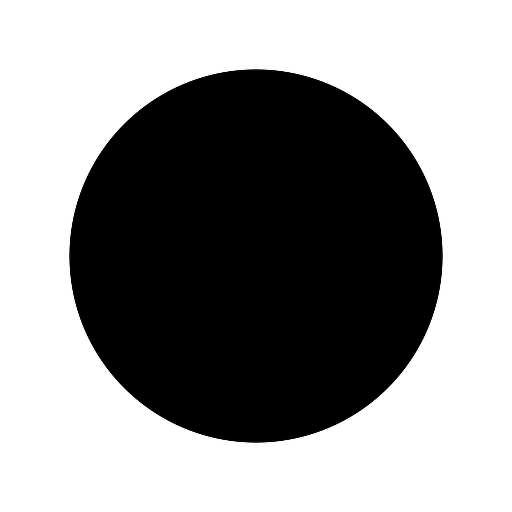}} &
            \fboxsep=0pt\fbox{\includegraphics[width=\itemwidth]{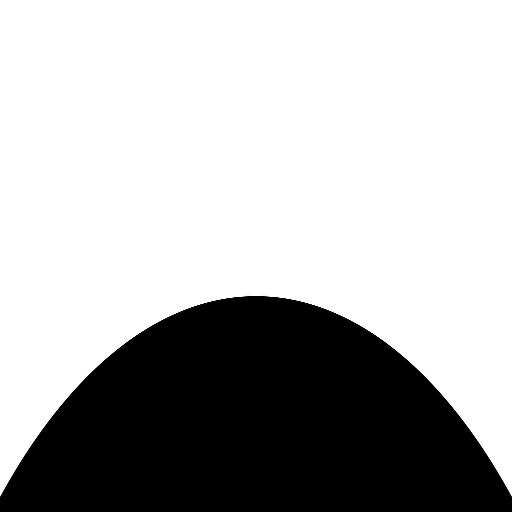}} &
            \fboxsep=0pt\fbox{\includegraphics[width=\itemwidth]{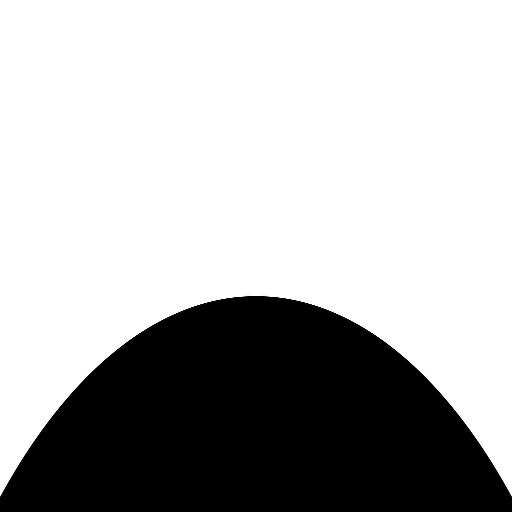}} \\\\
    \end{tabular}\vspace{-1em}
  \vspace{-0.25cm}\caption{\textbf{Masks for photographers.}}
  \label{fig:mask_photographer}
\end{figure}

To evaluate our approach, we introduce three urban scenes from the MovieMap~\cite{moviemap}. The MovieMap Dataset was created by sampling images from 360\textdegree~videos of varying lengths. Specifically, we extracted 51 images from a 3-second video, 12 images from a 6-second video, and 15 images from a 7-second video. Each image has a resolution of 3840$\times$1920. For each 360\textdegree~image, we extracted 14 perspective projection images. An example of this extraction process from a single 360\textdegree~image is illustrated in \Fref{fig:cut_out_pers}. A common challenge when capturing 360\textdegree~images is the inclusion of the photographer in the frame. To address this, we created a mask to exclude the photographer from the images, which is demonstrated in \Fref{fig:mask_photographer}. This masked area was subsequently omitted from both the training and evaluation phases.

\section{Additional Implementation Details}
\label{sec:additional_impl}
\subsection{Rendered Background-only Images of MovieMap Dataset}
When training static Neural Radiance Fields (NeRF) by removing masked objects, a significant challenge arises from errors near the edges of segmentation masks. These errors can disrupt the model's ability to accurately render static-only scenes, as they introduce inconsistencies at the boundaries of masked moving objects. To mitigate this, we dilated the mask area of moving objects. We implemented this by applying a convolution operation with a uniformly positive $3\times3$ kernel. Subsequently, in the output of this convolution, all positive values were converted to 1.

\subsection{Robust Approaches for Entity Segmentation Errors}
To prevent errors near the edges of entity segmentation, the area where the predicted entity-wise loss weights cover moving objects is increased through dilation. This is achieved by performing convolution with a uniform positive-valued $3\times3$ kernel and setting any positive values obtained in the result to 1.

Entity segmentation does not assign an entity to every pixel; some pixels are not assigned to any entity. Especially near the edges of objects, there are often pixels that were not assigned to any entity. We choose to include in training all pixels that are not classified as entities. However, we expect that the weight mask dilation process will exclude pixels near the edges of moving objects, which are not assigned to any entities, from the training process.

{
    \setlength{\tabcolsep}{2.5pt}
    \begin{table*}[t]
        \centering
        \small\begin{tabular}{c|rrr|rrr|rrr|rrr}
        & \multicolumn{3}{c|}{Statue} & \multicolumn{3}{c|}{Android} & \multicolumn{3}{c|}{Crab} & \multicolumn{3}{c}{BabyYoda} \\
        Loss & PSNR$\uparrow$ & SSIM$\uparrow$ & LPIPS$\downarrow$ & PSNR$\uparrow$ & SSIM$\uparrow$ & LPIPS$\downarrow$ & PSNR$\uparrow$ & SSIM$\uparrow$ & LPIPS$\downarrow$ & PSNR$\uparrow$ & SSIM$\uparrow$ & LPIPS$\downarrow$ \\\hline
        Mean-squared error (MSE) & 18.89 & 0.70 & 0.24 & 18.53 & 0.63 & 0.25 & 24.68 & 0.80 & 0.11 & 22.54 & \textbf{0.73} & \textbf{0.28} \\
        RobustNeRF~\cite{robustnerf} & 21.14 & \textbf{0.74} & 0.19 & 19.47 & 0.65 & 0.21 & 30.32 & 0.83 & \textbf{0.10} & 25.16 & 0.69 & 0.33 \\
        Entity-NeRF (only EARR) & 21.10 & \textbf{0.74} & \textbf{0.18} & 19.99 & \textbf{0.69} & \textbf{0.20} & 30.43 & 0.83 & \textbf{0.10} & 25.63 & 0.68 & 0.33 \\
        Entity-NeRF & \textbf{21.20} & 0.73 & 0.19 & \textbf{20.23} & 0.67 & 0.21 & \textbf{30.65} & \textbf{0.84} & 0.11 & \textbf{25.65} & 0.68 & 0.33 \\
        \end{tabular}
        \caption{\textbf{Quantitative comparison with RobustNeRF~\cite{robustnerf} using Mip-NeRF 360~\cite{barron2022mipnerf360} on RobustNeRF Dataset.}}
        \label{table:experiment_robustdata}
    \end{table*}
}

\section{More Results}
\label{sec:more_results}
\subsection{Evaluation on RobustNeRF Dataset}
\noindent\textbf{Dataset details}: Four natural scenes (i.e., Statue, Android, Crab, BabyYoda) from RobustNeRF~\cite{robustnerf}. Distractor objects are either moved or allowed to move between frames to simulate capture over extended periods. The number of unique distractors varies from 1 (Statue) to 150 (BabyYoda). Additional frames without distractors are provided to enable quantitative evaluation.

Note that we encountered issues with the provided camera parameters for the Statue and Android scenes, and the Crab scene does not provide camera parameters. Consequently, we calibrated the cameras using COLMAP~\cite{schoenberger2016sfm} for these scenes and used the calibrated parameters for training in the three scenes (Statue, Android, and Crab). The BabyYoda scene was trained using the original camera parameters.\\

\noindent\textbf{Quantitative comparison}: A quantitative evaluation using Mip-NeRF 360~\cite{barron2022mipnerf360} on the RobustNeRF natural scenes (Statue, Android, Crab, and BabyYoda), which were shot with objects centered, is shown in \Tref{table:experiment_robustdata}. Although our proposed method is not intended to improve the performance of scenes shot with the object centered, it showed that the proposed method outperformed RobustNeRF in PSNR, and was equal or better in terms of SSIM and LPIPS. Even when a moving object is photographed at a large size, the same problem as in the urban scene may occur because the object appears at the edge of the patch, and EARR appeared to have solved this problem. In addition, the incorporation of knowledge by the stationary entity classification was also found to be effective in the indoor scenes.

\begin{figure}[t]
  \centering
  \small{
  \begin{minipage}[t]{0.116\textwidth}
    \centering
    \includegraphics[width=\linewidth]{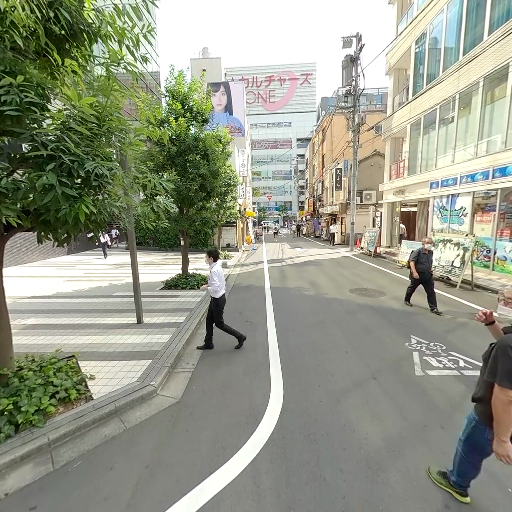}
    \vspace{-0.7cm}\center{Original image}
  \end{minipage}
  \begin{minipage}[t]{0.116\textwidth}
    \centering
    \includegraphics[width=\linewidth]{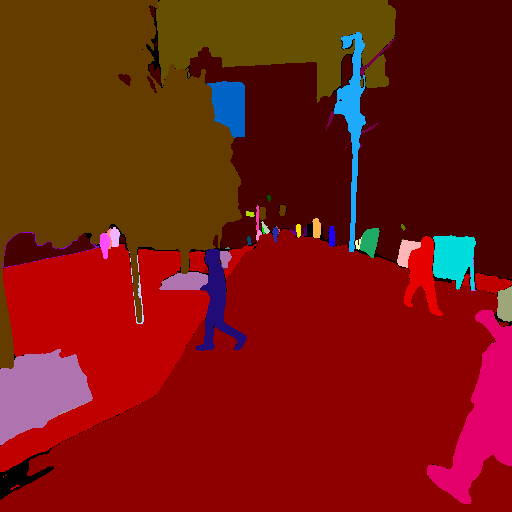}
    \vspace{-0.7cm}\center{Entity Seg.~\cite{qilu2023high}}
  \end{minipage}
  \begin{minipage}[t]{0.116\textwidth}
    \centering
    \includegraphics[width=\linewidth]{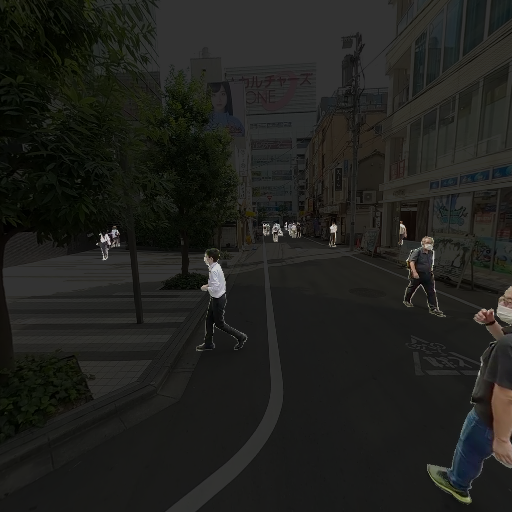}
    \vspace{-0.7cm}\center{Moving objects}
  \end{minipage}
  \begin{minipage}[t]{0.116\textwidth}
    \centering
    \includegraphics[width=\linewidth]{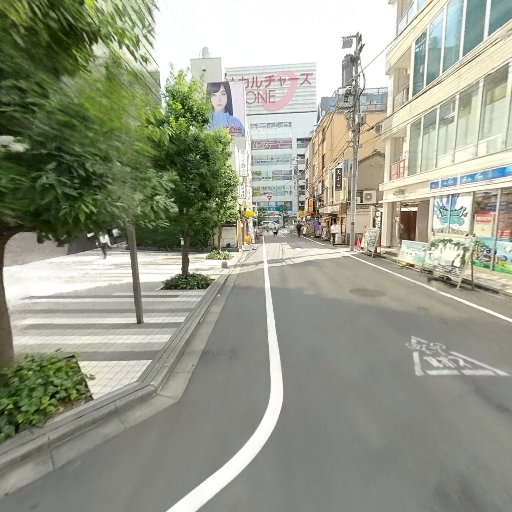}
    \vspace{-0.7cm}\center{Entity-NeRF}
  \end{minipage}
  \\
  \begin{minipage}[c]{0.116\textwidth}
    \centering
    \vspace{-1.5cm}\center{EARR}
  \end{minipage}
  \begin{minipage}[t]{0.116\textwidth}
    \centering
    \fboxsep=0pt\fbox{\includegraphics[width=\linewidth]{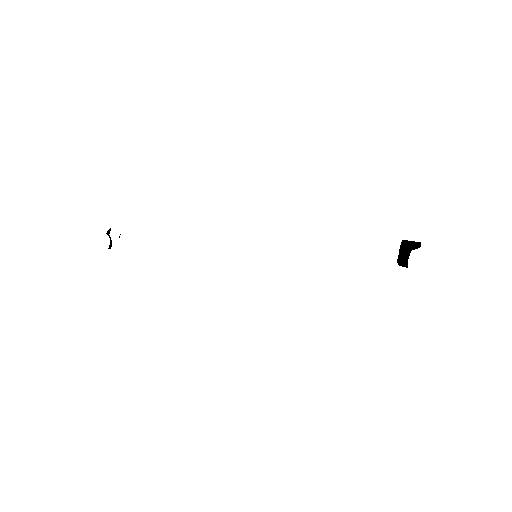}}
  \end{minipage}
  \begin{minipage}[t]{0.116\textwidth}
    \centering
    \fboxsep=0pt\fbox{\includegraphics[width=\linewidth]{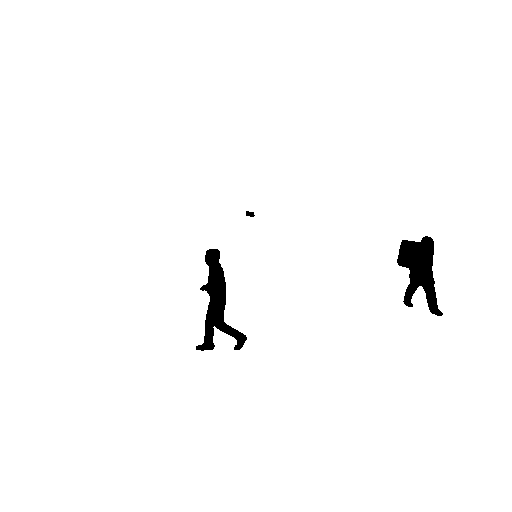}}
  \end{minipage}
  \begin{minipage}[t]{0.116\textwidth}
    \centering
    \fboxsep=0pt\fbox{\includegraphics[width=\linewidth]{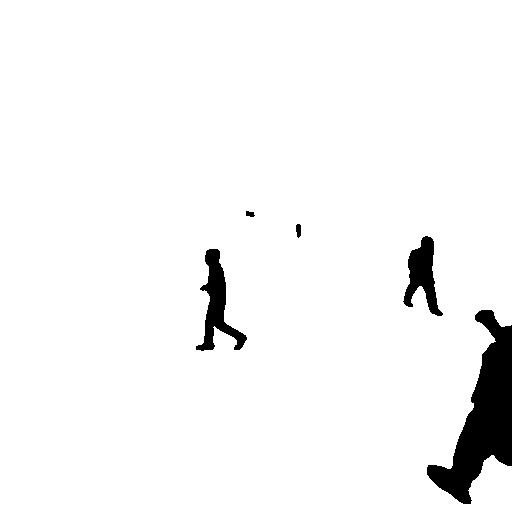}}
  \end{minipage}
  \\
  \begin{minipage}[c]{0.116\textwidth}
    \centering
    \vspace{-1.7cm}\center{Stationary Entity Classification}
  \end{minipage}
  \begin{minipage}[t]{0.116\textwidth}
    \centering
    \fboxsep=0pt\fbox{\includegraphics[width=\linewidth]{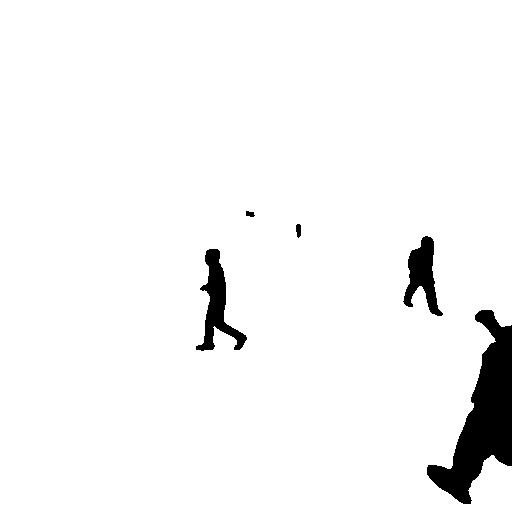}}
    \vspace{-0.7cm}\center{500 iterations}
  \end{minipage}
  \begin{minipage}[t]{0.116\textwidth}
    \centering
    \fboxsep=0pt\fbox{\includegraphics[width=\linewidth]{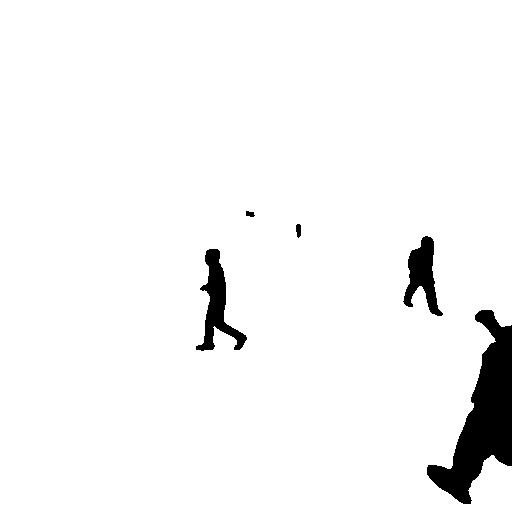}}
    \vspace{-0.7cm}\center{1,000 iterations}
  \end{minipage}
  \begin{minipage}[t]{0.116\textwidth}
    \centering
    \fboxsep=0pt\fbox{\includegraphics[width=\linewidth]{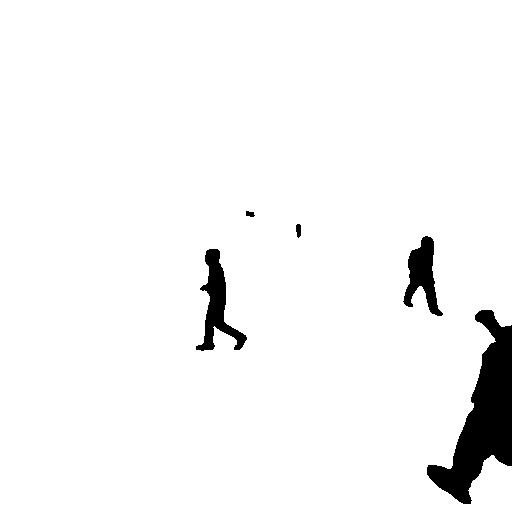}}
    \vspace{-0.7cm}\center{5,000 iterations}
  \end{minipage}
  }
  \vspace{-0.25cm}\caption{\textbf{Visualization of $\bm{D(}\mathbf{r}\bm{)}$ using stationary entity classification.} Compared to EARR, \(D(\mathbf{r})\) in the early stages of training are improved.}
\label{fig:vis_neural_weight}
\end{figure}

\subsection{Qualitative Comparison of Distractiveness using stationary entity classification}
As shown in \Fref{fig:vis_neural_weight}, our thing/stuff segmentation-based stationary entity classification provides more precise \(D(\mathbf{r})\) assignments for each entity than EARR in initial learning stages, where predicting accurate diffuse \(D(\mathbf{r})\) for all entities is challenging due to large residuals.

\subsection{Quantitative Comparison of Distractiveness}
In \Tref{table:dataset_details}, the Intersection over Union (IoU) of Distractiveness $D(\mathbf{r})$ in Entity-NeRF at the end of training is compared with the IoU of Distractiveness $D(\mathbf{r})$ in RobustNeRF~\cite{robustnerf}, using masks annotated on moving objects as ground-truth labels. Our proposed method achieves a better IoU for both $D(\mathbf{r})=0$ and $D(\mathbf{r})=1$, allowing for closer Distractiveness to the annotated mask to be given as a weight in the loss.

\begin{table}[t]
    \centering
    \small\begin{tabular}{c|cc}
        & \begin{tabular}
    {c}IoU\\$D(\mathbf{r}) = 1\uparrow$\end{tabular} & \begin{tabular}
    {c}IoU\\$D(\mathbf{r}) = 0\uparrow$\end{tabular} \\\hline
        RobustNeRF~\cite{robustnerf} & 0.84 & 0.14 \\
        Entity-NeRF & \textbf{0.98} & \textbf{0.59} \\
    \end{tabular}
    \caption{\textbf{Quantitative Comparison of Distractiveness with RobustNeRF~\cite{robustnerf} on MovieMap Dataset}}
    \label{table:dataset_details}
\end{table}

\subsection{Novel View Synthesis}
\begin{figure}[t]
    \centering
    \setlength{\tabcolsep}{0.02cm}
    \setlength{\itemwidth}{2.7cm}
    \renewcommand{\arraystretch}{0.5}
    \hspace*{-\tabcolsep}\begin{tabular}{ccc}
            \includegraphics[width=\itemwidth]{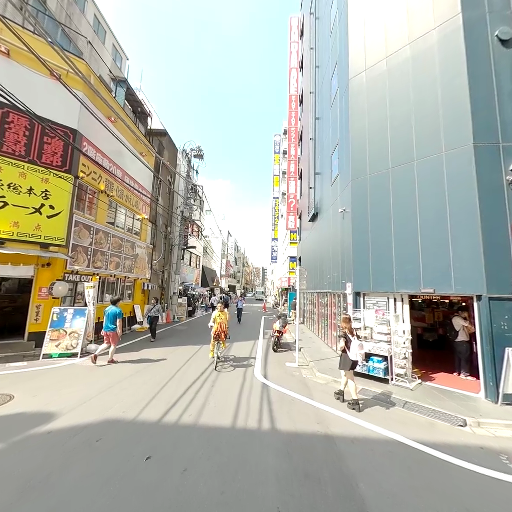} &
            \includegraphics[width=\itemwidth]{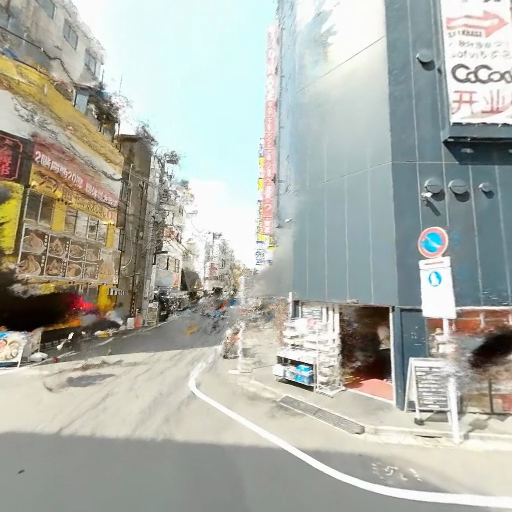} &
            \includegraphics[width=\itemwidth]{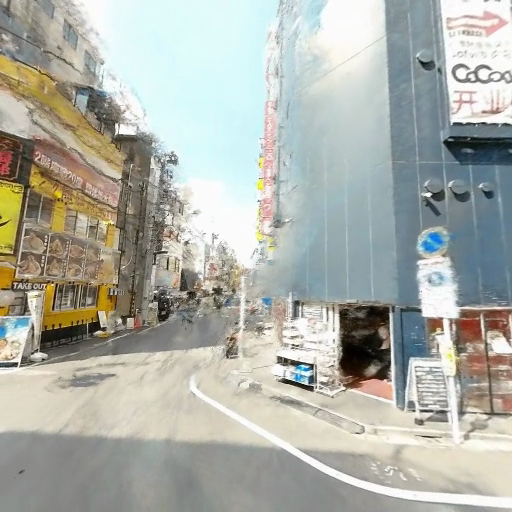} \\
            \includegraphics[width=\itemwidth]{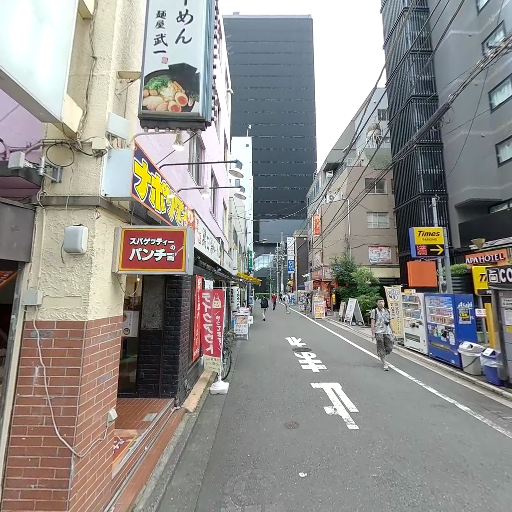} &
            \includegraphics[width=\itemwidth]{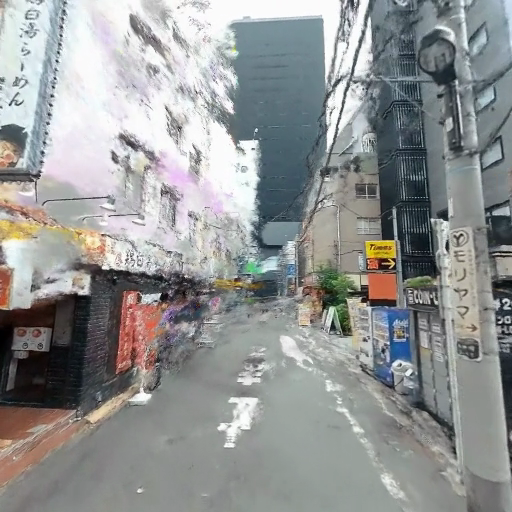} &
            \includegraphics[width=\itemwidth]{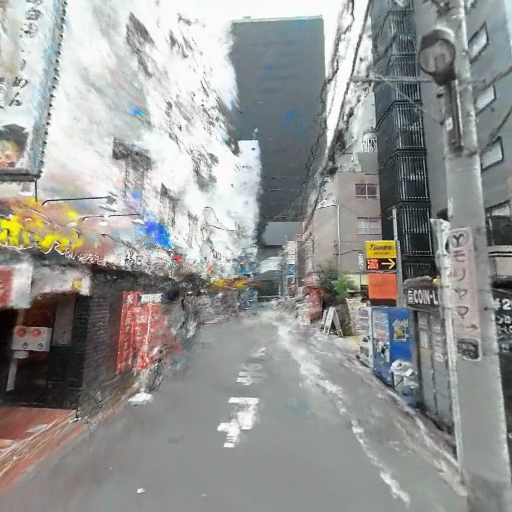} \\
            \includegraphics[width=\itemwidth]{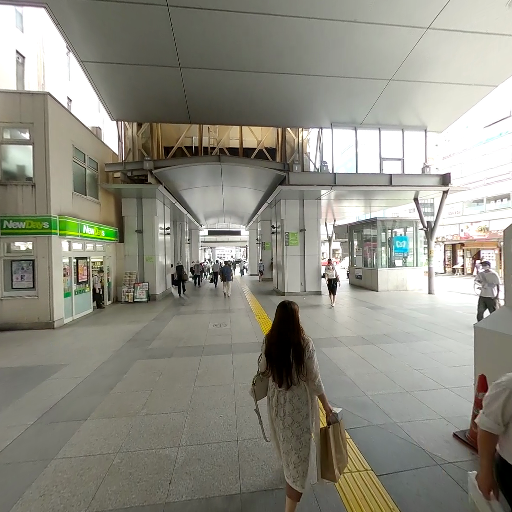} &
            \includegraphics[width=\itemwidth]{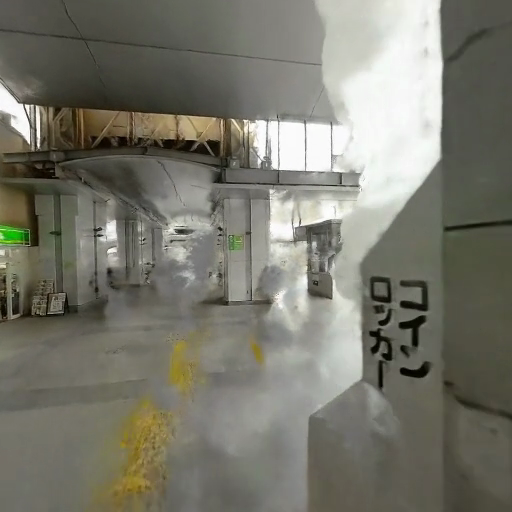} &
            \includegraphics[width=\itemwidth]{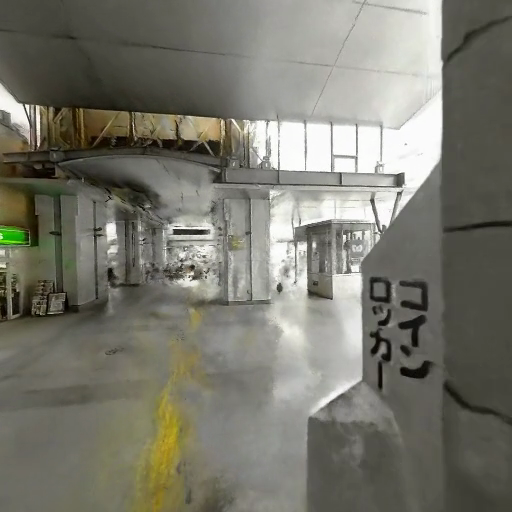} \\
            Reference image & RobustNeRF & Entity-NeRF\\
    \end{tabular}
  \vspace{-0.25cm}\caption{\textbf{Novel view synthesis.}}

  \label{fig:nvs}
\end{figure}

We conduct a qualitative comparison of our Entity-NeRF's performance in novel view synthesis. The novel view synthesis using the MovieMap Dataset is shown in \Fref{fig:nvs}. We created a circular trajectory around the straight-line path of the original video and synthesized new views on the circular path. Entity-NeRF, although suffering from degradation due to the inability to learn correct geometry, shows less deterioration compared to RobustNeRF~\cite{robustnerf}. This is evident from the comparison of synthesized images from different viewpoints during training, as our approach avoids erroneously including moving objects in the training process and includes more static backgrounds into the training.

